\documentclass[conf]{new-aiaa}
\usepackage[utf8]{inputenc}

\usepackage{graphicx}
\usepackage{multirow}
\usepackage{amsmath}
\usepackage{bm}
\usepackage[version=4]{mhchem}
\usepackage{siunitx}
\usepackage{longtable,tabularx}
\usepackage{booktabs}  % for table lines
\usepackage{multirow}  % for spanning rows

\usepackage{fancyhdr}% for temporary
\usepackage[absolute,overlay]{textpos} % for textblock* environment

\usepackage{algorithm}
\usepackage{algpseudocode}

\usepackage{comment}

\usepackage{xcolor}

\usepackage{ulem}

  {\endgroup}                % End: close group

\title{Ground Plane-Aided Extrinsic Calibration of Inertial and RGB-D Sensors for Uncrewed Aerial Vehicles}

\author{Ilyar Asl Sabbaghian Hokmabadi\footnote{Postdoctoral Associate, Department of Mechanical and Manufacturing Engineering, 40 Research Pl NW, Calgary} and Mahdis Bisheban \footnote{Assistant Professor, Department of Mechanical and Manufacturing Engineering, 40 Research Pl NW, Calgary, AIAA Member. This work was supported by the Natural Sciences and Engineering Research Council of Canada (NSERC), the Government of Alberta, Alberta Innovates, and the Schulich School of Engineering at the University of Calgary. Funding was awarded to Dr. Mahdis Bisheban, Director of the Intelligent Dynamics and Control Lab and Assistant Professor at the University of Calgary.}}
\affil{University of Calgary, Calgary, Alberta, AB T2L 1Y6}

\begin{document}

\begin{textblock*}{\textwidth}(1cm,1cm)
\centering
\small Author's preprint. Published by AIAA SciTech Forum, 2026. URL = {https://arc.aiaa.org/doi/abs/10.2514/6.2026-1068}
\end{textblock*}

    \maketitle
    
%%Temporay:
%\pagestyle{fancy}
%\fancyhead{} % clear all header fields
%\fancyhead[RO,LE]{\textbf{Accepted for Publication in the 2026 AIAA SciTech}}
%%end Temporary

    \begin{abstract}
            Accurate extrinsic calibration of inertial sensors, such as Inertial Measurement Units (IMUs) and cameras is crucial for trajectory estimation of Uncrewed Aerial Vehicles (UAVs). While numerous calibration methods have been proposed, these techniques often rely on specialized equipment, planar targets, and an initial estimate of the calibration parameters. 
            In this research, we propose a targetless calibration method designed for UAVs equipped with IMUs and RGB-Depth (RGB-D) cameras. Our approach leverages deep-learning-based floor-segmentation to extract ground points from the depth channel of RGB-D images. Subsequently, the normal vector to these points is estimated. The known orientation of the normal to the floor segment and the gravity vector sensed in the accelerometer's frame are utilized in a robust estimation approach to estimate the extrinsic calibration parameters. We illustrate that the developed method outperforms MATLAB's Toolboxes and exhibits similar performance to Kalibr without the use of specialized checkerboard targets.

            %The proposed floor segmentation achieves up to $0.96$ mean precision and recall. It is also shown that including IMU intrinsic calibration results in an approximate $2.65^\circ$ improvement in the estimated extrinsic orientation. 
    \end{abstract}
        	
    \section{Introduction}     
    Sensor fusion between Inertial Measurement Units (IMUs) and cameras is critical for autonomous systems. This fusion has been widely used to estimate the attitude, position, and velocity of Uncrewed Aerial Vehicles (UAVs) \cite{ollero2021past}. However, accurate path and trajectory estimation depends on the precise determination of extrinsic calibration parameters, which are defined as the translation and rotation between the coordinate frames of the sensors.
    
% 2. categorization of extrinsic calibration and the issues with the current ones,
    Extrinsic calibration methods can be categorized into target-based and targetless approaches. Target-based methods \cite{rehder2016extending,zhang2021monocular} rely on specialized equipment such as checkerboards.  
    A key limitation of these methods is the need to maintain visibility of the calibration target, which is particularly challenging in UAV applications.
    Target-less calibration methods \cite{qin2018vins,huang2020online,huai2022observability,wang2021adaptive} rely on online trajectory estimation. 
    However, this approach can be unreliable due to the need for the availability of visual features, which depend on environmental texture and lighting conditions. 
    This can be challenging as camera-based trajectory estimation relies on salient point detection and matching in consecutive images. Such salient points can only be detected if there are objects in the environment with visual textures, under suitable illumination conditions. Moreover, low-cost IMU-based trajectory estimation can also be challenging due to error accumulation. Therefore, it is important to develop calibration methods that do not depend on salient feature point detection and special visual targets. Further, such calibration should be independent of the trajectory estimation of a UAV unlike the methods proposed in \cite{lee2022extrinsic,ouyang2025dynamic}.%\CMB{Can you cite some examples of methods depending on trajectory?}ILyar: Although, other methods mentioned in the paragraph rely on the trajectory estimation, I found two studies with also an explicit mention of the "trajectory" even in their title. 
    %\CMB{How calibration can be dependent on or independent of the trajectory estimation of a UAV?}

    %Extrinsic calibration parameter initialization is crucial in many methods \cite{yang2016monocular}. %
    The initialization problem for extrinsic calibration parameters is crucial and recognized as challenging \cite{yang2016monocular}. 
    In \cite{yang2016monocular}, the authors address this initialization problem by first estimating the relative orientation between two camera frames at consecutive epochs using Nistér’s five-point algorithm \cite{nister2004efficient}. The relative orientation of the IMU between two frames captured at two consecutive epochs  in the trajectory is also estimated using gyroscope measurements. The initial extrinsic calibration is then obtained by minimizing the error between these two relative rotations. However, this approach still relies on two consecutive frames. \cite{xiao2025slope} proposes a coarse initialization method for LiDAR–IMU calibration. Their method requires a sloped surface to estimate the relative extrinsic orientation between the IMU and the LiDAR. This approach is similar to \cite{yang2016monocular} in that it also requires estimating the relative motion between the two frames. In contrast, the method proposed in this paper utilizes corresponding accelerometer–RGB-D measurement directly, without relying on two-view relative motion estimation, therefore, reducing the computational cost of the algorithm.
    
    These limitations highlight the need for calibration methods that do not rely on artificial targets or continuous trajectory estimation. Our research addresses this gap by proposing a robust, targetless calibration approach that is independent of trajectory estimation and better suited for UAV operations. 
    In our method, the floor normal vector is assumed to be parallel to the gravity vector, which is a reasonable assumption in structured, indoor environments. 
    By aligning the estimated floor normal vector with the gravity vector measured in the accelerometers' coordinate frame, the relative orientation between the IMU and RGB-D camera can be inferred. 
% 3. Introduction to the proposed method, briefly
    While the proposed method relies on the visibility of the ground segment in RGB-D images, this is a reasonable assumption in most indoor, human-made environments, where flat floors are typically unobstructed. Unlike traditional feature-based methods that depend on textured objects or rich visual scenes, our approach leverages the structural regularity of indoor floors, which are consistently present and detectable even under varying lighting conditions. Unlike methods \cite{rehder2016extending} that require continuous feature detection, the proposed calibration approach uses discrete pairs of IMU measurements and RGB-D images containing visible ground plane segments to estimate the extrinsic parameters.
    
    % 4. Introduction to the utilization of the ground plane for land vehicles
      Ground plane detection plays a key role in many land-based robotic applications. For land vehicles with a fixed monocular camera height relative to the ground, the scale ambiguity of visual features on the ground is immediately resolved using the available height information. Ground plane detection helps identify these important features and assigns 3D coordinates to them \cite{Kitt-2011-7357} , without requiring feature matching and multi-frame triangulation, and ultimately improves camera trajectory estimation \cite{zhou2019ground_monoscale}. Ground plane detection is also important in imposing height constraint on LiDAR/camera during the trajectory estimation of wheeled vehicles. This height constraint is crucial for the ubiquitous LiDAR-based odometry. LiDAR height obtained from LiDAR odometry is often not restricted to a direction parallel to the ground. This is due to the fact that features such as trees or street poles do not provide a strong vertical constraint. This will lead to error accumulation in the vertical (to the ground plane) direction. In order to address this challenge, ground-plane detection is utilized to identify features lying on this plane and help constraining the estimated height of the trajectory \cite{wei2025enhanced_GC_LOAM}. In \cite{kim2024gril_wheeledground} ground plane detection is utilized to constrain the height estimate of a mobile wheeled robot, as well as to estimate the extrinsic calibration parameters of the LiDAR and IMU. \cite{ai2023lidar} utilized ground-plane height constraint similar to \cite{kim2024gril_wheeledground} and improved the accuracy further by utilizing map matching between the current LiDAR point cloud and the available pre-processed point cloud of the environment. Ground-plane detection has also been utilized for RGB-D-based motion estimation of mobile wheeled robots in \cite{li2025lg_wheeledground}. %Ilyar: The height constraint is a constraint in the optimization sense. Such constraints are valid for land-vehicle trajectory estimation. If sucha  constraint is not explicitly included in the trajectory estimation, the errors in the vertical direction can grow. Specifically for land-vehicle LiDAR odometry, many objects, such as poles and tree trunks, do not provide a strong constraint in the vertical direction.
    
    % 5. Introduction to the utilization of the ground plane for an aerial vehicle
    Ground-plane detection plays a key role in the navigation and control of a UAV, as well. \cite{pritzl2023adaptive} uses a downward-looking rangefinder to detect the ground and estimate the height of a UAV in the indoor environment. However, since a UAV can hover over objects above the ground, such height estimation is prone to errors. \cite{zhao2025visual} proposes a method to estimate the ground plane in order to estimate the constant height of a UAV in a cluttered under-canopy environment. More recently, in \cite{whitney2025global}, ground plane detection has been employed to allow for the reduction of the dimensionality of 3D pose estimation to 2D and improve the computational cost of particle filtering. %\CMB{Can you clarify the reduction of the dimensionality of 3D pose estimation to 2D ?} 
    Their method reduces the particle sampling space dimensions by first estimating the height of the UAV from the ground. The estimated height restricts particle filter sampling of the position from 3D to a 2D plane.
    Ground-plane detection is also useful in emergency landing for UAVs. \cite{chen2021integrated_emergencylanding} proposes a deep learning based image segmentation method to identify the largest empty space on the ground for a UAV to land in an emergency landing situation.

    % 6. Introduction to the ground-plane detection (classic/not learning based)
    Because of the importance of ground/floor detection, many researchers have focused on improving its accuracy. In \cite{berenguel2020floor_visuallyimpaired}, a ground/floor segmentation approach is proposed using two cameras. A downward-looking RGB-D camera is utilized to initiate the seeds for a region-growing algorithm. The corresponding pixels in a fish-eye camera (the second camera) are found and expanded to detect the entire floor segmentation. This approach assumes the availability of a downward-looking RGB-D camera. 
    The authors in \cite{tian2021accurate} detect the ground plane by first estimating the trajectory of the monocular camera. The estimated trajectory and the corresponding identified features in the images are triangulated. Subsequently, the normal to each point is extracted by fitting local planes. They assumed that the ground plane normal is orthogonal to the direction of the camera's motion throughout the vehicle's trajectory, which is not an assumption made in our proposed method. %\CMB{The following paragraph is about LIDARs, and it does not seem to be relevant to this paper? Is it? Ilyar: The depth channel can be converted to a point cloud, which is the type of data received from the LiDAR, and there are many commonalities between the algorithms developed.  Based on my understanding, the depth in and RGB-D can be estimated using different technologies. We have depth from stereo, which relies on two cameras, depth from structured light, and depth from Time of Flight. The last technology shares the working principles with some of the LiDARs. What is sold as LiDAR typically has a longer range and a rotating part inside, which increases the resolution. I added a sentence in the end that, although these LiDARs with longer ranges can help, they do not help with being heavier!
    Ground plane detection can be facilitated by using sensors with rich 3D information, like 3D LiDARs. A survey of LiDAR-based ground plane detection \cite{gomes2023survey} shows that this plane can be estimated using LiDAR data. This is especially true for ground vehicles and mobile wheeled robots, as typically, the approximated height of the LiDAR from the ground is known. In \cite{liu2024flexible_GC_detection}, the authors propose to cluster LiDAR points into height intervals. This limits the search for the ground plane to a certain interval, reducing the computational cost. However, relying on the assumption of constant sensor height is not applicable in many UAV-based applications. Further, LiDARs are more expensive and heavier, making them less suitable for low payload UAV missions compared to RGB-D cameras. The proposed method achieves floor-segmentation with the help of the RGB channel, instead.
    
  % 7. Introduction to the ground-plane detection (learning based)  
    Classical methods, introduced above, do not rely on data-driven, learning-based algorithms for ground detection and segmentation. They are not robust because ground and specifically indoor floor planes are often made of reflective material, which produces unpredictable features for classical hand-crafted feature-detection approaches such as Scale Invariant Feature Transform (SIFT) \cite{lowe2004distinctive}, thereby limiting the broader applicability of these methods. More recently, learning-based methods have been explored for floor and ground segmentation \cite{man2019groundnet}. In \cite{man2019groundnet}, a deep-learning-based approach for ground plane segmentation and surface-normal detection is proposed; however, their method is primarily tested on road segments. Similarly, \cite{hamandi2018ground} proposed a fully convolutional neural network (FCN)-based approach comparable to the method presented in this paper. However, their model is trained for outdoor ground detection, such as garden beds, whereas the proposed method is trained for indoor environments. In \cite{paigwar2020gndnet}, the ground plane is detected by first projecting the 3D point cloud onto a 2D image-like grid, where point height is encoded in the feature channel, and a neural network is trained on these 2D representations for ground segmentation. The proposed method of this paper does not require the depth channel for ground detection and relies only on the RGB information.

     %\CMB{ A very short literature review on the normal vector to the floor plane detection methods.}
     Floor-normal detection is an important step in the proposed extrinsic calibration method. The normal of the floor plane is obtained after floor segmentation. In particular, the depth values of pixels belonging to the floor segment can be used to estimate the floor-plane normal. Since RGB-D measurements are contaminated with noise, robust plane-fitting methods are required. Plane fitting to noisy point clouds can be achieved using methods such as the Hough transform, originally introduced in \cite{duda1972use}. Hough transformation is a histogram-voting algorithm in which a stencil of lines passing through each point is parameterized by orientation and distance to the origin. Votes are then added to the corresponding bins for each point. The bin with the highest number of votes is selected as the solution. The Hough transform is computationally feasible in 2D because the histogram involves only two parameters; however, extending it to 3D plane fitting increases the dimensionality, which significantly increases computational complexity. This high computational cost in higher-dimensional spaces is examined in \cite{li2023integrated}. Another method for robust plane fitting is DBSCAN \cite{schubert2017dbscan}, a point cloud density-based clustering algorithm. Its robustness makes it suitable for rejecting outlier points before fitting a plane to the inliers. DBSCAN classifies data points into three groups: core, reachable, and noise points. Core points have a minimum number of neighboring points within a specified radius. Reachable points lie within this radius of a core point. Points that do not satisfy either condition are labeled as noise. DBSCAN requires tuning of two hyperparameters. RANSAC is another robust plane-fitting technique that removes outliers during estimation. Many variants of RANSAC have been introduced since its inception \cite{martinez_otzeta2022ransac}, and RANSAC has been used for plane fitting \cite{li2017improved}. Like DBSCAN, RANSAC requires hyperparameter tuning. Finally, Singular Value Decomposition (SVD) can also be used to fit a plane. SVD is a computationally efficient method and is used for datasets containing relatively few outliers. Due to the high accuracy of the floor segmentation in our pipeline, we rely on SVD for floor-plane fitting to reduce computational cost.

    We summarize the contributions of this work as follows: (1) A targetless extrinsic calibration method for an RGB-D camera and IMU is proposed. The developed method leverages semantic segmentation to extract the floor segment from RGB-D camera images. The normal of this plane and the gravity vector from the IMU are used to estimate the extrinsic calibration parameters. (2) The proposed method does not require an initialization of the extrinsic calibration parameters. (3) The developed method is a general-purpose extrinsic calibration approach for an IMU and RGB-D camera and can be used on both land and aerial robots. No assumption about a fixed RGB-D camera relative to the ground plane is made(4) We further demonstrate the importance of intrinsic IMU calibration for achieving more accurate extrinsic calibration. 

  \section{Methodology}\label{section:methodology}

\begin{figure}
            \centering
            \includegraphics[
                width=\textwidth,
                height=\textheight,
                keepaspectratio]{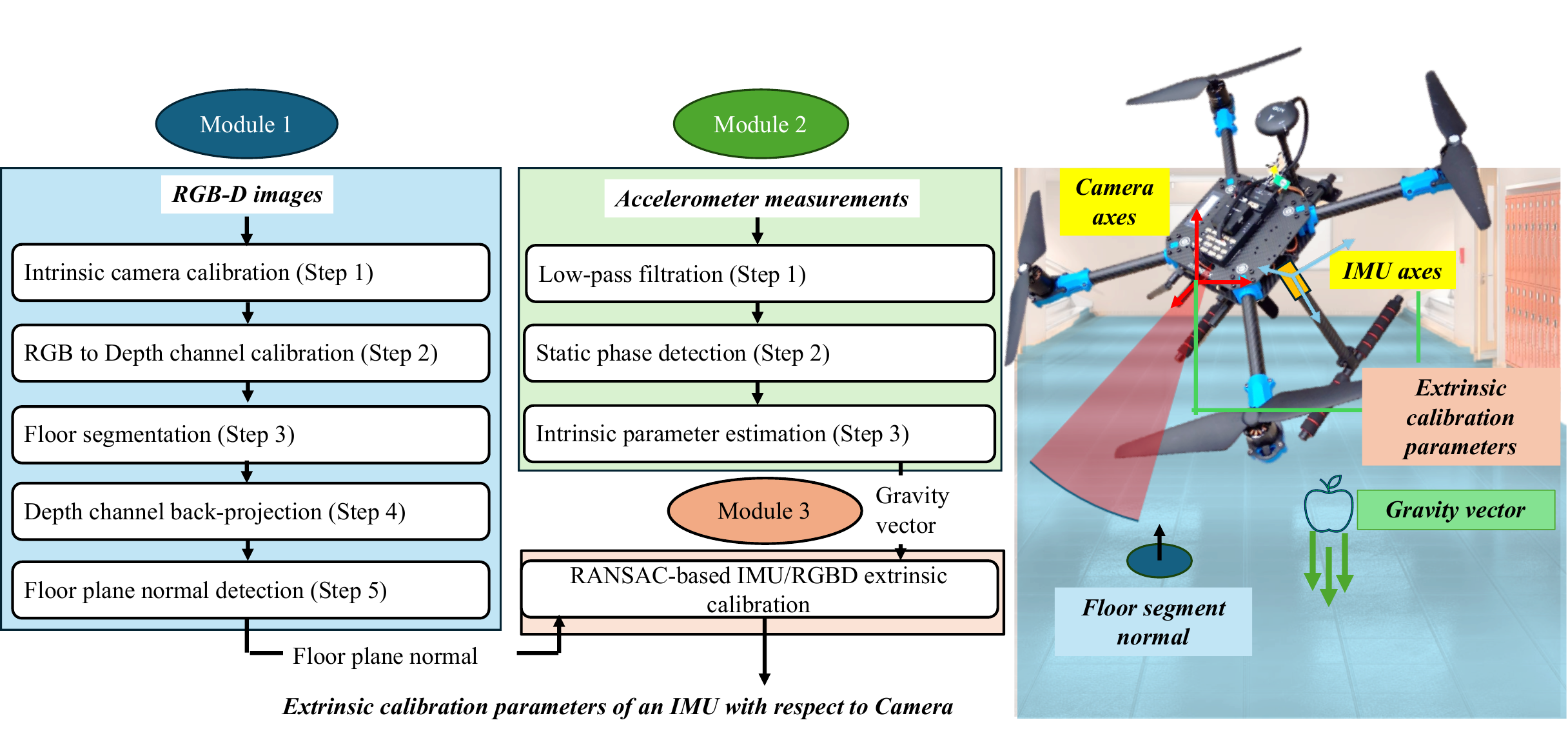}
            \caption{The proposed extrinsic calibration of an IMU and an RGB-D camera for UAV-based trajectory estimation}
            \label{fig:1}
    \end{figure}

  The developed extrinsic calibration approach consists of three modules. The first two modules perform RGB-D image processing and accelerometer processing, explained in Section \ref{section:methodologyA} and \ref{section:methodologyB}, respectively. The detected floor normal and the gravity vector are then matched to estimate the extrinsic orientation parameters, explained in Section \ref{section:methodology_c}. An overview of the proposed method is shown in Fig. \ref{fig:1}.
    
    \subsection{Module 1: RGB-D Image Processing}\label{section:methodologyA}

    The first module includes five steps: pinhole camera calibration, RGB to depth channel calibration, floor segmentation, depth channel pixel back-projection, and floor plane normal detection. 
    
    Pinhole camera calibration (Module 1, Step 1) includes estimation of the location of the orthogonal projection of the optical center of the camera onto the image plane. This projection is often close to the center of the image and is denoted with two coordinates as $c_x \in \mathbb{R}$ and $c_y \in \mathbb{R}$. Pinhole camera calibration also includes the focal length, which is the distance from the optical center to the image-forming plane. The focal lengths in the $x$ and $y$ axes ($f_x \in \mathbb{R}$, $f_y \in \mathbb{R}$) are considered separate parameters. The fifth pinhole camera calibration parameter is the skew parameter $\gamma \in R$.
    In total, pinhole camera calibration includes five parameters ($f_x, f_y, c_x, c_y, \gamma$). For most cameras, these parameters are only required to be estimated once. The pinhole camera calibration matrix is shown in Equation \eqref{eq:camera_intrinsic}, where $\bar{u}$ and $\bar{v}$ denote the normalized coordinates of the pixels and ${u}$ and ${v}$ are the pixels in the image frame. 
    \begin{equation}\label{eq:camera_intrinsic}
    \begin{bmatrix}
    u \\ v \\ 1
    \end{bmatrix}
    =
    \begin{bmatrix}
    f_x & \gamma & c_x \\
    0   & f_y & c_y \\
    0   & 0   & 1
    \end{bmatrix}
    \begin{bmatrix}
    \bar{u} \\ \bar{v} \\ 1
    \end{bmatrix}.
    \end{equation}

    Pinhole camera calibration parameters cannot capture non-linear distortions. Such non-linear distortions can cause straight lines to appear curved in images. The non-linear distortions are typically included in the intrinsic camera calibration process as well. The most significant non-linear distortion is known as radial distortion, which happens due to the opening and closing of the camera shutter. Assuming that the distorted normalized coordinates of the pixels are denoted as $\bar{u}_{dist}$ and $\bar{v}_{dist}$, and radial distortion coefficients are denoted as $\lambda_1$, $\lambda_2$, and $\lambda_3$, radial distortion equations are shown in the following \cite{zhang1999flexible},
    \begin{equation}
    \begin{aligned}
    \bar{u}_{dist}  = \bar{u} \left( 1 + \lambda_1 r^2 + \lambda_2 r^4 + \lambda_3 r^6 \right),\\
    \bar{v}_{dist} = \bar{v} \left( 1 + \lambda_1 r^2 + \lambda_2 r^4 + \lambda_3 r^6 \right),
    \end{aligned}
    \end{equation} where $r^2 = \bar{u}^2 + \bar{v}^2$. In our experiments, we have used MATLAB\texttrademark\ Camera Calibration Toolbox \cite{camera_calibration_doc} and a checkerboard in an offline process to estimate these parameters. 

    %Ilyar: Intrinsic camera calibration includes pinhole parameters. It can further include radial coefficients as well, if the  camera exhibits severe Barrel distortions. I have now included the radial distortions equations as well. Further, I modified Figure 1 to intrinsic calibration rather than just pinhole.
    Followed by the intrinsic camera calibration, the unknown translation and the orientation  of the depth channel, the camera should be estimated (Module 1, step 2). This step ensures that there is a correct alignment between the pixels in the depth channel and the pixels in the RGB channel. These parameter (which includes a rotation $R_{RGB}^{D} \in SO(3)$ and the translation vector $T_{RGB}^{D} \in \mathbb{R}^3$) are provided by the manufacturer. %Ilyar: T and R will be provided, but in the Appendix as we agreed upon previously. Although necessary, we did not derive these parameters ourselves; they do not belong in the results section. However, this is a crucial step if such parameters are not provided by the manufacturer, and it must be mentioned in the flowchart and as a Step.
    
    In Module 1, Step 3, the images obtained from the RGB channels of the camera are utilized to segment the floor region. In this work, in order to ensure that floor segmentation is achieved with high accuracy, a deep neural network is trained to segment the floor in RGB images. Once RGB-based floor segmentation is achieved, the corresponding pixel depth can be extracted from the depth (D) channel. The  deep neural network utilized for this training is a fully convolutional neural network (FCN), which has an encoder-decoder architecture described.
    %in Table \ref{tab:architecture}. 
    The input size to this neural network is equal to height ($H$) and width ($W$) of the images. The output is a binary image with the same height and width as the input image. The corresponding pixels to the floor segment are labeled as binary high (one), and the other pixels are labeled as binary low (zero). The trained neural network has four encoder and decoder layers. The input to this network is an image of 256 by 256 RGB pixels. The convolutional kernel size is 2 by 2 with a kernel stride of 2.
    %\CMB{iss pooling layers? convolutional layers typically use 3×3 kernels}. 
    The network includes 31,031,745 parameters in total, and it is trained using approximately 100,000 images. The database includes the flipping, rotation, contrast variation, and blurring data augmentations. The loss function is defined as the binary cross-entropy. Adam's optimizer with an initial learning rate of 0.001 with a batch size of 16 is utilized. %The initial weights are not set to any specific values. 
    The output of a binary segmentation algorithm requires a threshold to identify the foreground pixels (the floor segment) from the background pixels. In our experiments, this value is 0.5 (the value range of the output pixels is between 0 to 1).

    In Step 4, the depth values corresponding to the segmented floor pixels are back-projected using the camera intrinsic matrix to generate a 3D  point cloud ${(x, y, z) \in R^3}$ in the camera coordinate frame, where each point represents a physical location on the floor surface.
    In order to back-project these pixels, the pinhole camera model (outcome of Module 1, Step 1), and the RGB to depth channel calibration (outcome of Module 1, Step 3) are required. With these, the following back-projection equation can be utilized 
    \begin{equation}\label{eq:backproject}
    \begin{bmatrix}
    x \\
    y \\
    z
    \end{bmatrix}
    =
    Z \;
    \begin{bmatrix}
    \frac{1}{f_x} & -\frac{\gamma}{f_x f_y} & \frac{\gamma c_y - c_x f_y}{f_x f_y} \\
    0 & \frac{1}{f_y} & -\frac{c_y}{f_y} \\
    0 & 0 & 1
    \end{bmatrix}
    \begin{bmatrix}
    u^{F} \\
    v^{F} \\
    1
    \end{bmatrix},
    \end{equation}
    where $Z \in \mathbb{R}$ is the normalized ($[0,1]$) gray-scale pixel value of the depth channel, and $u^{F}$ and $v^{F}$ are the coordinates of the pixels residing on the floor-segment. It is important to note that a global scale to transform the back-projected pixels to metric scale is required. This scale is constant for all the pixels, and for an RGB-D camera, it is often provided by the manufacturer.
    %(this value be \CMB{?} tuned for finer or coarse depth precision \CMB{?}). \CMB{Provide this value in the result section.} 
    
     Next, in Module 1, Step 5, the floor plane normal is detected. In most environments, particularly indoor spaces, the floor segment corresponds to a set of co-planar points that lie on a common plane. These points, extracted from the depth channel of an RGB-D image, represent the visible portion of the floor surface. However, depending on the environment, the co-planarity assumption might not always hold. The normal vector to the segmented plane can be estimated using well-known plane fitting algorithms (Module 1, Step 4). In the proposed method, Singular Value Decomposition (SVD) is utilized to estimate a best-fit normal vector to these points. SVD algorithm outputs singular vectors. These vectors are ordered such that the first component corresponds to the direction with the largest variance, and the last component corresponds to the direction with the smallest variation in the data. 
     Since points corresponding to a planner surface are expected to vary the least in the direction perpendicular to the plane, the vector corresponding to the smallest singular value is considered as the plane normal. In order to show the derivation of the floor normal in the camera frame, mathematically, we commence by stacking the homogeneous coordinate points of the point cloud in a matrix $P_k \in \mathbb{R}^{m \times 4}$ shown in the following, where $m$ and $k$ are the number of points and the number of current epoch, respectively.
     \begin{equation} \label{eq:stacked_coordinates}
        P_{k} = 
        \begin{bmatrix}
        x_1 & y_1 & z_1 & 1 \\
        x_2 & y_2 & z_2 & 1 \\
        \vdots & \vdots & \vdots & \vdots \\
        x_m & y_m & z_m & 1
        \end{bmatrix}.
    \end{equation}

       %SVD can be used as $P_k = U \, D \, V^\top$. %shown in Equation \eqref{eq:SVD_lidar_points}. 
       The SVD algorithm decomposes the matrix $P_{k} = U \, D \, V^\top$ to three matrices $U \in \mathbb{R}^{3\times 3}$, $D\in \mathbb{R}^{3\times 3}$ and $V\in \mathbb{R}^{3\times 3}$, where $U$ and $V$ are orthonormal matrices. % such that ($U^TU = I, V^TV = I$). 
       $D$ is diagonal with positive real values.
    %\begin{equation} \label{eq:SVD_lidar_points}
        %P_k = U \, D \, V^\top.
    %\end{equation}
    With the help of SVD, the smallest eigen vector ($V(:,4)$) can be selected. This eigen vector corresponds to the floor normal of the point cloud, and it is shown in the following,
\begin{equation} V(:,4) = 
\begin{bmatrix}
d_{1} \\ d_{2} \\ d_{3} \\ d_{4}
\end{bmatrix}, 
\quad \text{with the plane equation defined as} \quad d_{1} x + d_{2}  y + d_{3} z + d_{4} = 0.
\end{equation}

Finally, the plane normal $v^{C}_{k} \in \mathbb{R}^3$ , which is required in Module 3, can be calculated by normalizing the $V(:,4)$ using the homogeneous part of the vector in the following, where the superscript $c$ denotes the current camera frame,

\begin{equation}\label{eq:normalizd_floor_normal}
v^{C}_{k} = \frac{1}{\sqrt{d_{1}^2 + d_{2}^2 + d_{3}^2}} 
\begin{bmatrix} d_{1} \\ d_{2} \\ d_{3} \end{bmatrix}.
\end{equation}

    %UNet \cite{ronneberger2015u} architecture%,
   %Ilyar comment: I actually made a mistake and did not include the skip connections in the new implementation. Currently,y it is just an encoder-decoder network, and I corrected this in the text to reflect this.
   
%Ilyar comment: It seems that the row is correct, as the input is the point cloud and the output is the floor normal. And yes, SVD is used to extract that normal vector from the point cloud. I will add all the details of this part if time permits. But it is SVD-based. 
 
%Ilyar comment: For RGB-D camera calibration itself, which we have on the Github page, I was extracting the scale using the area from the checkerboard  images, and that is why that algorithm was more complicated. Here, I assumed this scale is roughly available. Adding this step will divert the paper from the main goal, which is the extrinsic calibration, I thought. Maybe I should include a phrase that such a global scale is available.
   
    \subsection{Module 2: Accelerometer Processing}\label{section:methodologyB}
    
    Module 2 includes three steps. Low-pass filtering of accelerometer measurements, static phase detection, and estimation of intrinsic parameters. The output of this module is an estimated of the gravity vector. 
    Low-pass filtering (Module 2, Step 1) is the first step that ensures the removal of high-frequency components of the signal's spectrum. In the experiments, Butterworth filtering is utilized with a cut-off frequency of $f_c$ and an order of $L$. 
    The frequency filtering is applied to each axis of accelerometer data independently. 
    
    Followed by this low-pass filtering, in Step 2, the static phases are detected using the magnitude of the accelerometer data. A static phase is an interval where the magnitude of change in the accelerometer's data is smaller than a certain threshold. Such a threshold is sensor dependent, and in this work, the value is set to 0.01 ($m/s^2$). Further, a minimum window size is used to reject false data in the static phase detection. The size of this window depends on the time the sensor is kept in static phases. In our experiment, the window size corresponds to approximately $0.5$ seconds. The importance of utilizing the static phase detection is due to the fact that intrinsic accelerometer calibration (Module 2, Step 3) relies on known measurements of the gravity vector's magnitude. The magnitude of the accelerometer readings can only be assumed to be equal to gravity when no other external forces are applied to this sensor. These phases here are denoted as static phases.
     
     %In the third step, intrinsic calibration of the IMU and the gyroscope is estimated with the help of the gravity vector magnitude.
    
   % \SMBilyar{To complete the extrinsic calibration process, the gravity vector must be estimated using IMU measurements (\EMB{module 2, step 5}).  During static phases, the accelerometer measures only the gravity. Such static phases occur when the sensor is not in motion (e.g., placed on a table). However, this assumption does not hold when the UAV is in motion. During motion, the gravity vector in the static phase can be tracked using gyroscope-based integration, assuming the initial gravity is sensed using the accelerometer.}% Iyar comment: This part is completely modified in the implementation. While static phase detection is still important for the intrinsic calibration of the accelerometer, since gyroscopes are not used, no integration is necessary. Further, I found out that the data can be used in the estimation of the extrinsic calibration parameters even in the dynamic phases.

    % 4 & Estimating gravity vector & Calibrated accelerometer ($\boldsymbol{a}$) & Gravity vector in the IMU frame ($g$) & Accelerometer readings \\

      In Step 3, the intrinsic calibration of accelerometers is estimated. This step is important to track the direction of gravity accurately using accelerometers. Intrinsic calibration helps mitigate the errors between the expected and the actual outputs of an IMU. These parameters are independent of how the IMU is mounted on the UAV. For a triaxial accelerometer, these calibration parameters can be summarized as scale ($\boldsymbol{s} \in \mathbb{R}^3$), bias ($\boldsymbol{b} \in \mathbb{R}^3$), and non-orthogonality ($\boldsymbol{n} \in \mathbb{R}^3$). The bias is defined as an offset from the zero measurements when no external forces are applied to the sensor (the accelerations are zero). The scale factor is the ratio of the true and the actual outputs of the sensor. Non-orthogonality captures any projection of the three axes of a triaxial accelerometer on other axes. Ideally, for a sensor, these axes should be mutually orthogonal with non-orthogonal values of zero.
      
      In order to calibrate the accelerometer, the method in \cite{tedaldi2014robust,khankalantary2021simplification} is utilized. This approach requires a UAV or the device to be placed at different static phases. Static phases are utilized to estimate all the unknown linear calibration parameters of an accelerometer (Step 3). These calibrated accelerometers measure the gravity vector in the sensor's body frame, which can be used directly in the estimation of the extrinsic calibration as it will be explained in Section \ref{section:methodology_c}. This process does not require gyroscopes. This is an important advantage as gyroscope measurements can suffer from g-dependent bias variations \cite{wang2021g_sensitivity} and, in general, bias instability \cite{qureshi2017infield}, making its calibration challenging. Further, utilizing this approach, instruments such as a triaxial turntable are not required; thus rendering the proposed approach suitable for in-suite calibration. The calibration model used for an accelerometer (Module 2, Step 3) is shown in the following:
        \begin{equation}\label{eq:IMU_intrinsic}
            \boldsymbol{a}_{k}^{B} = M \tilde{\boldsymbol{a}}_{k}^{B} + \boldsymbol{b}_{k} + \boldsymbol{\eta},  \quad \boldsymbol{\eta} \sim \mathcal{N}(0, \Sigma).
        \end{equation}
        where $\tilde{\boldsymbol{a}} \in \mathbb{R} ^3$ denotes uncalibrated and filtered accelerometer measurements, while $\boldsymbol{a} \in \mathbb{R}^3$ refers to the calibrated and filtered accelerometer measurements. Random errors in the measurements are modeled as a Gaussian distribution shown as ($\boldsymbol{\eta} \in \mathbb{R}^3$) with the covariance matrix shown as $\Sigma \in \mathbb{R}^{3 \times 3}$. The superscript $B$ indicates that the measurements are obtained in the sensor body frame. These measurements are utilized to estimate the unknown orientation of the camera with respect to the body frame.
        The calibration matrix ${M} \in \mathbb{R}^{3 \times 3}$ includes axes non-orthogonality ($\boldsymbol{n}=[n_{xy},n_{xz},n_{yz}]$), and the scale factor ($\boldsymbol{s}=[s_x,s_y,s_z]$) as follows 
        \begin{equation}\label{eq:calibmat_expanded}
             M =\begin{bmatrix}
            s_{x} & n_{xy} & n_{xz}\\
            0 & s_{y} & n_{yz}\\
            0 & 0 & s_{z}
            \end{bmatrix}.
        \end{equation}    

    %Module 2 is summarized in Table \ref{tab:accelerometer_modules}.

    \subsection{Module 3: RGBD/IMU Extrinsic Calibration}\label{section:methodology_c}
     % \CMB{Please explain the trajectories. For example all axes of IMUs need to be excited?}
     Module 3 is the proposed Ground-plane based RANdom SAmple Consensus (RANSAC) IMU/RGB-D extrinsic calibration, where unknown extrinsic calibration parameters of the camera and the IMU are estimated by aligning the floor-normal vector and gravity vector over $K$ samples. The unknown extrinsic calibration parameters of the camera and the IMU are shown by $R_{B}^{C} \in SO(3)$. The $K$ samples are gathered over specific trajectories. During these trajectories, IMUs should experience different poses with respect to the gravity vector. This ensures that the extrinsic calibration can be achieved successfully. Specifically, the device (with IMU and camera) should experience rotations around three perpendicular axes.
     
     In Module 3, the proposed method estimates a rotation matrix (corresponding to the unknown extrinsic orientation) that can minimize the total error between the gravity vector in the IMU and RGB-D frames as shown in the following,
        \begin{equation}\label{eq:error term}
            e= \frac{1}{2}\sum_{k}{w_k||{\boldsymbol{v}_k^C-\ R_B^C\ {\boldsymbol{a}}_k^B||^2}},
        \end{equation} 
        where $\boldsymbol{v}_{k}^C \in \mathbb{R}^3 $ is the floor-plane normal in the RGB-D camera frame, ${\boldsymbol{a}}_k^B \in \mathbb{R}^3$ is the calibrated acceleration in the static phase in IMU frame, which roughly corresponds to the gravity vector if the sensor is static or it is not moved too fast. The filtering of the accelerometer introduced in the previous section helps remove high-frequency vibrations from the data. The unknown $R_{B}^{C} \in SO(3) $ is the extrinsic calibration matrix, which $w_{k} \in \mathbb{R}^K$ is a weight coefficient. 
        %\CMB{how to differentiate gravity with a general reading? (${\boldsymbol{a}}_k^B \in \mathbb{R}^3$ is the gravity vectors in IMU frame)%Ilyar: The measurement from acceleration corresponds to gravity in the static phase, as no other forces are being applied to it.}
        Equation \eqref{eq:error term} is known as Wahba’s problem \cite{wahba1965least}. There are numerous approaches proposed to address this problem \cite{wu2021lasso_wahba,chen2024fracgm}. In this paper, we propose a RANSAC-based solution. First, we reformulate Equation \eqref{eq:error term} as a standard least squares estimation. In order to achieve this, we commence from the following,
    \begin{equation}\label{eq:towardsLS}
        \boldsymbol{v}^{C}_{k} = R^{C}_{B}  {\boldsymbol{a}}_k^B.
    \end{equation}
    % \EMB{where $\boldsymbol{v}^{c}_{k}$ and ${\boldsymbol{a}}_k^B$ are the gravity vectors in camera and IMU frames respectively, and $R^{B}_{c}$ is the unknown rotation matrix between the coordinate frame of the camera and the IMU.}
    
    We can rewrite Equation \eqref{eq:towardsLS} in terms of 9 unknown parameters of the rotation matrix as shown in the following,
    \begin{gather} \label{eq:rewritten_towardLS}
        \boldsymbol{v}^{c}_{k} = 
        \begin{bmatrix}
        {}^{x}a_{k} & {}^{y}a_{k} & {}^{z}a_{k}  & \boldsymbol{0}_{1 \times 3} & \boldsymbol{0}_{1 \times 3} \\
        \boldsymbol{0}_{1 \times 3} & {}^{x}a_{k} & {}^{y}a_{k} & {}^{z}a_{k}& \boldsymbol{0}_{1 \times 3} \\
         \boldsymbol{0}_{1 \times 3} & \boldsymbol{0}_{1 \times 3} & {}^{x}a_{k} & {}^{y}a_{k} & {}^{z}a_{k}
        \end{bmatrix} \boldsymbol{r} ,
    \end{gather}
     where the bold symbol $\boldsymbol{r} \in \mathbb{R}^9$ indicates vectorized nine unknown parameters of the extrinsic rotation matrix. The three equations shown above represent one measurement correspondence at a time $k$. The left-hand side and the matrix rows can be stacked as for $k \in [1,K]$, as shown in the following 
    \begin{equation}\label{stacked_towardsLS}
        \boldsymbol{v}^{c}_{1:K}  = A \boldsymbol{r}.
    \end{equation}
    %\EMB{where $K$ is the number of data points. Each data point includes 
    %one image and one measurement of IMU..
    The solution to this standard least square form is shown in the following
    \begin{equation} \label{eq:solved_LS}
        \boldsymbol{r} = (A^{T} W^{-1} A) ^{-1} W ^{-1} A^{T} \boldsymbol{v}^{c}_{1:K},
    \end{equation}
    where $W \in \mathbb{R}^ {3K \times 3K}$ is the weight of the measurements (similar to the Wahba's formulation).
    %\CMB{report the weight in the result section.}

    The estimated rotation parameters between an IMU and a camera can be reorganized into a matrix form, $\tilde{R}^{C}_{B}\in \mathbb{R}^{3\times 3}$. In addition, the orthogonality constraint can be imposed using SVD. The estimated rotation parameters can be reorganized into a matrix form ($\tilde{R}^{C}_{B}$), then we can compute the SVD of this matrix as 
    %\begin{equation}\label{eq:SVD_R}
    $ \tilde{R}^{C}_{B} =  U D V^{T} $.
    %\end{equation}
    Finally, multiplying $U$ and $V^{T}$ will result in a corrected rotation matrix ${R}^{B}_{C}$, as shown in 
    \begin{equation} \label{eq:SVD_R_corrected}
        {R}^{C}_{B} = U V^{T}.
    \end{equation}
    To validate the proper rotation, we ensure the determinant is $+1$ and $R^T R =I$.

    Removing outliers is an important step, as some of the floor-normal and gravity measurements in the data can have errors that cannot be mitigated relying only on the least square-based technique shown above. Such robustness can be achieved by randomly sampling from the measurements to find an inlier set. The proposed method requires three measurement pairs. Each measurement pair provides three equations, and there are in total of nine equations to estimate the unknowns. Once the extrinsic calibration parameters are estimated, they are tested using residual errors of the other measurements. Outliers are defined as those measurements with residual error larger than a user-defined threshold $\tau$. RANSAC stops when an extrinsic calibration with a sufficiently large inlier-to-outlier ratio is found. The pseudo code of the developed RANSAC-based robust estimation is shown in Algorithm \ref{alg:RANSAC}.

    \begin{algorithm}
    
        \caption{RANSAC \cite{martinez_otzeta2022ransac} for Extrinsic Calibration using Floor Normals and Gravity Vectors}
        \label{alg:RANSAC}
        \begin{algorithmic}[1]
        \Require  Floor normals from RGB-D camera ${\boldsymbol{v}}_k^c$, gravity vectors from IMU $ {\boldsymbol{a}}_k^B$, threshold $\tau$
        \Ensure Estimated rotation matrix ${R}^{B}_{c}$
        \State Initialize: Inlier threshold $\tau$, best rotation ${R}^{B}_{c} \gets {I}$
        
        \While {$\alpha > \tau$}
        
            \State Randomly select a minimal subset of correspondences: $({\boldsymbol{v}}_k^c, {\boldsymbol{a}}_k^B)$
            
            \State Estimate rotation ${R}^{B}_{c}$ using \eqref{eq:solved_LS}-\eqref{eq:SVD_R_corrected}
            
            \State Initialize inlier count $\alpha \gets 0$
            \For{Remaining pairs of ${\boldsymbol{v}}_k^c$,$ {\boldsymbol{a}}_k^B$ }
            
                \State Compute $ error : ||{{\boldsymbol{a}}}_k^B-\ R_c^B\ \boldsymbol{v}_k^c|| $ 
                
                \If{ $error < \tau$}
                    \State $\alpha \gets \alpha + 1$
                \EndIf
                
            \EndFor
        \EndWhile
        
        \State \Return   ${R}^{B}_{c}$ 
        
    \end{algorithmic}
    \end{algorithm}

   % The steps above conclude with the third and final steps of the proposed calibration method (Step 3 in Fig. \ref{fig:1}). 
   
   %It is important to note that UAV-based Kalman filtering \cite{ramos2023reef} can be utilized to integrate and use the readings of an accelerometer to continuously update the estimated attitude during flight as well. This approach does not require multiple static phases as proposed in this paper. However, it is susceptible to error accumulation and can lead to the lowering of the overall accuracy of the system.

\begin{comment}
Module 3 is summarized in Table \ref{tab:fusion_modules}.
\begin{table}[!htbp]
\centering
\caption{Module 3: IMU-RGB-D Fusion and Extrinsic Calibration}
\renewcommand{\arraystretch}{1.3}
\begin{tabular}{|p{4.5cm}|p{4.5cm}|p{4.5cm}|}
\hline
  \textbf{Input} & \textbf{Output} & \textbf{Method} \\
\hline
 Floor plane normal ($\boldsymbol{v}_k^c$), Gravity vector ($\boldsymbol{a}_k^B$) & IMU-to-camera extrinsic parameters (Rotation matrix $\tilde{R}^{c}_{B}$) & Algorithm \ref{alg:RANSAC} \\
\hline
\end{tabular}
\label{tab:fusion_modules}
\end{table}
\end{comment}

    \section{Results and Discussion}\label{section:results} 

   % \CMB{Please provide the result of "each and every step" described in the methodology section here. This includes 5 steps of Module 1, and three steps of Module 2, and the outcome of Module 3.}
   %Ilyar: Module 1-1 is intrinsic calibration which I reported the results. Module 1-2 is the extrinsic of RGB to depth that needs to be reported from the sheet as discussed, Module 1-3 is the floor-segmentation, which we show an example and report precision/recall. For modules 1-4 and 1-5, I added figures projecting the normal vector of the floor onto the image. 
    %Ilyar: Module 3 is reported as well
    %Ilyar: For Module 2-1 is the filtering for this module I decided to show the results of the extrinsic calibration for different cut-off frequencies in the end. I did not show the filtered data as this is  a bit too basic, as you mentioned. But the final results show the extrinsic calibration for different values of cut-off frequency.
    %Module 2-2 and 2-3 are both reported adequately.
   
    % Module 1
    % Step 1 intrinsic camera calibration (check)
    % Step 2 RGB to depth channel calibration (check?manufacturer, so we did not calculate anything as our results)
    % Step 3 Floor segmentation (Check)
    % Step 4 Depth channel back-projection ()
    % Step 5 floor plane normal detection (check)
    
    %Our preliminary results show improved accuracy compared to other techniques such as FLAE \cite{wu2017fast_quat}, QUEST \cite{cheng2014improvement_quest}, weighted least squares (WLS), and it is summarized in Table \ref{tab:1}. 
        
\begin{figure}[!b]
    \centering
    \includegraphics[width=0.5\linewidth]{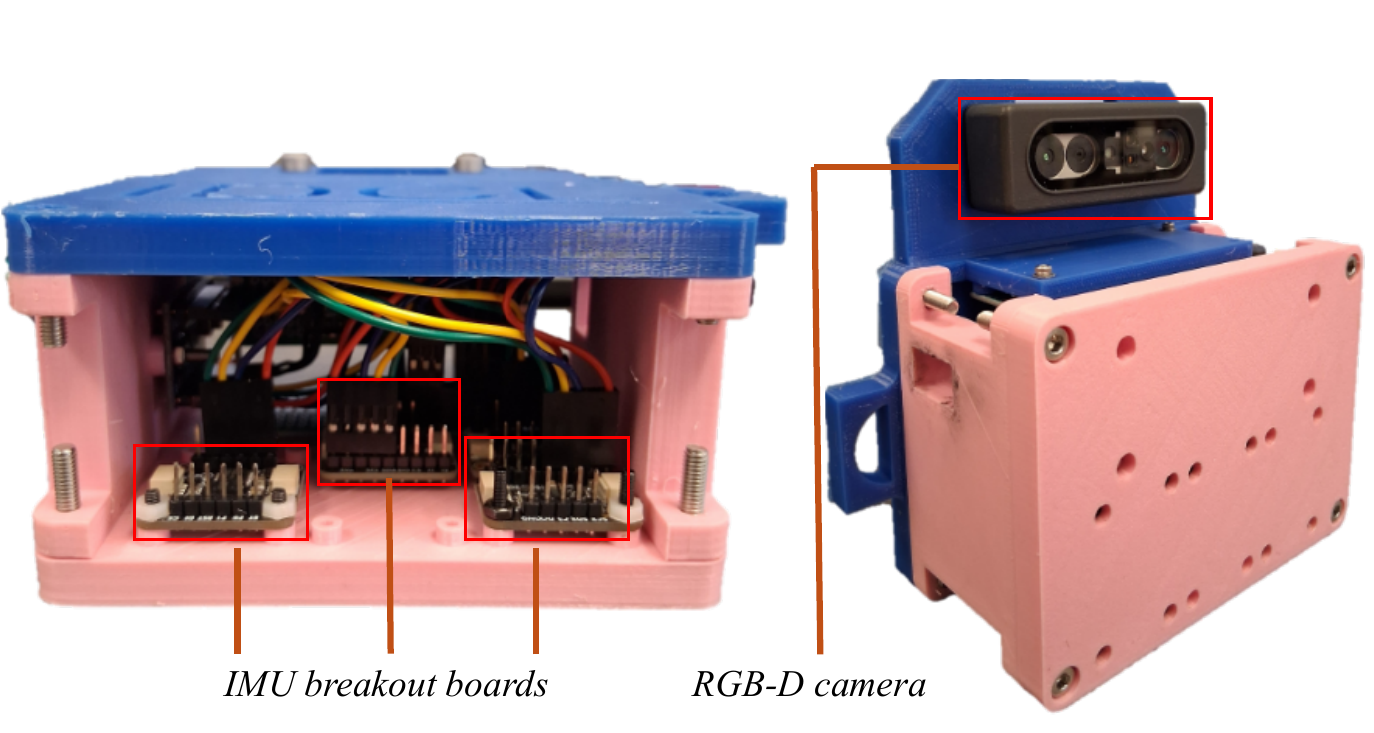}
    \caption{Experimental platform showing front and side views with an RGB-D camera and three MEMS IMUs}
    \label{fig:placeholder}
    \end{figure}

    To validate the developed method, low-cost sensors are used.
    These sensors are mounted on a platform as shown in Fig. \ref{fig:placeholder}. The sensors are connected to an onboard computer where the incoming measurements are time-synchronized. 
    Specifically, we used the Gemini 2 RGB-D camera by Orbbec\texttrademark{} \cite{orbbecGemini2}. This RGB-D camera is equipped with one stereo pair and an IR distance sensor. %, very similar to the architecture of Microsoft\texttrademark{} Kinect 2 \cite{microsoftKinectV2}. 
    This camera measures the distance at a range of 0.25 to 2.5 meters with a horizontal field of view of $67.9^\circ$ and a vertical field of view of $45.3^\circ$. The depth accuracy of the sensor at 1 meter is approximately 5 millimeters. This camera outputs RGB and depth. The depth channel is aligned to one of the stereo cameras using the manufacturer's calibration parameters. For this research, images of 640 $\times$ 480 pixels are utilized. The camera frame rate is set to $5 Hz$.
    
    Three IMUs, ISM330DHCX \cite{adafruitISM330DHCX}, LSM6DSOX \cite{adafruitLSM6DSOX} by STM\texttrademark{} , and MPU6050 \cite{invensenseMPU6050} by TDK\texttrademark, are used in the tests. 
    %ISM330DHCX and LSM6DSOX are developed by STM\texttrademark{} and MPU6050 is developed by TDK\texttrademark. 
    Expansion electrical boards are designed by Adafruit Industries LLC\texttrademark{} for these IMUs. 
    The IMUs are from different generations to ensure the applicability of the developed calibration method across different generations of MEMS IMUs. MPU6050 is from an older generation of sensors. % that are widely utilized in many research projects. The 
    LSM and ISM series belong to the newer generation. The input data rate for these IMUs was set to $60 Hz$. The input data rate for MPU6050 was set to $80 Hz$.  
 
    \begin{figure}[!b]
                \centering
                \includegraphics[
                width=\textwidth,
                height=\textheight,
                keepaspectratio]{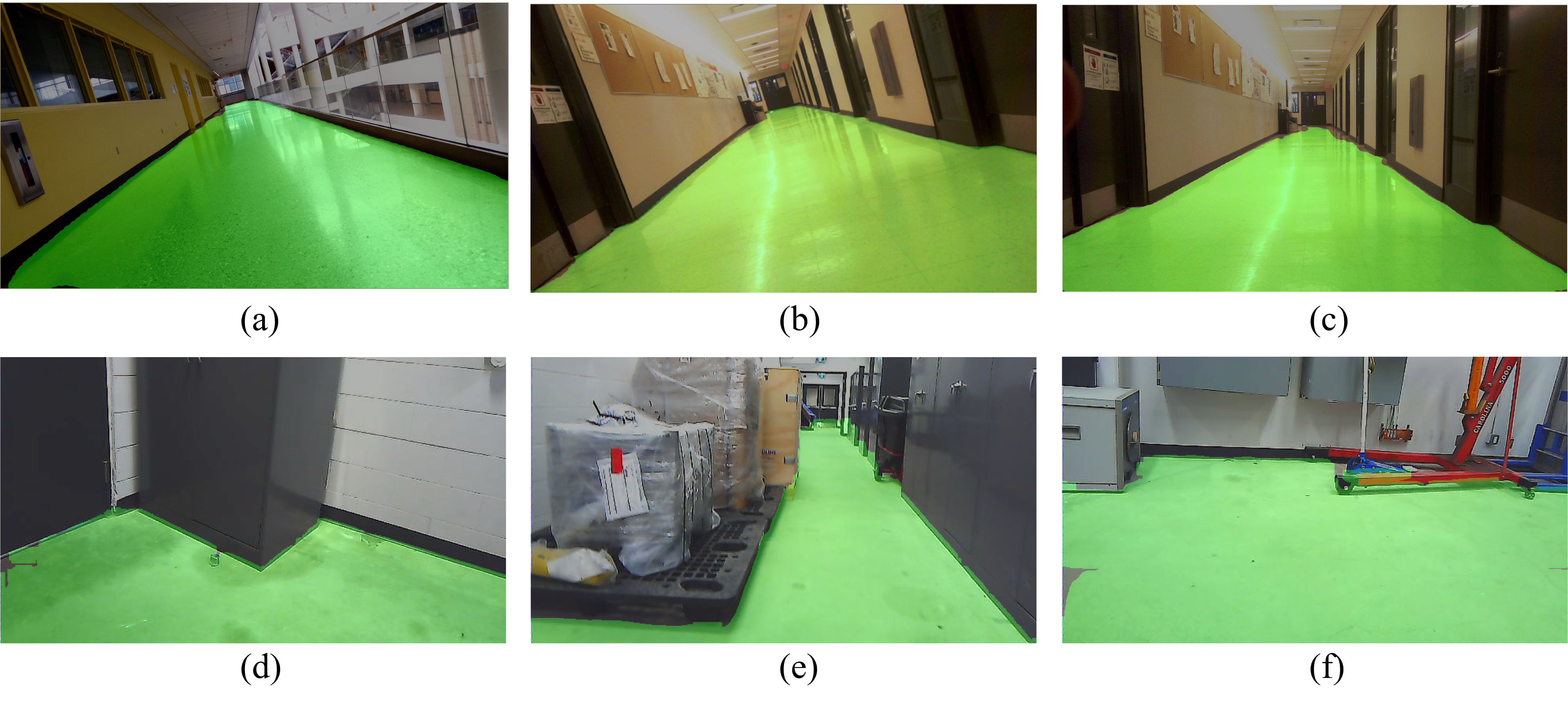}
            \caption{Floor detection results across different indoor environments (green: detected floor)} 
            \label{fig:floor_segment_examples}
    \end{figure}

   % report Module 1, Step 1 and Step 2
   %Ilyar: Module 1 step 1 is reported, and we can add Module 1 step 2 to the supplementary. 

    The camera intrinsic calibration parameters of Module 1, Step 1 are reported in Table \ref{tab:intrinsic_camera_calibration}. In total, 19 images are used to estimate these parameters. The mean reprojection error achieved is sub-pixel. This indicates a high quality of calibration. 
    Besides the pinhole camera model, radial distortion was found in the image as well. The estimated radial distortions ($\lambda_1, \lambda_2, \lambda_3$) are also reported in Table \ref{tab:intrinsic_camera_calibration}.
    RGB to depth calibration parameters in Module 1, Step 2 ($R_{RGB}^{D}$ and $T_{RGB}^{D}$) are provided by the manufacturers. 

    \begin{table}[!htbp]
    \centering
        \caption{Estimated camera intrinsic parameters}
    \begin{tabular}{c|c}
    \hline
    Calibration parameter & Value (unit) \\
    \hline
    $f_x$   & 455.1313 (mm/pixels) \\
    $f_y$   & 453.6879 (mm/pixels) \\
    $c_x$   & 338.1614 (pixels)\\
    $c_y$   & 241.9856 (pixels) \\
    $\gamma$   & -0.6977 (unitless)\\
    $\lambda_1$   & 0.079  (unitless)\\
    $\lambda_2$   & -0.042 (unitless)\\
    $\lambda_3$   & -0.163 (unitless) \\
    Mean reprojection error of checkerboard corners & 0.20 (pixels)\\
    Checkerboard images count & 19\\
    \hline
    \end{tabular}
    \label{tab:intrinsic_camera_calibration}
    \end{table}

\begin{figure}
        \centering
        \includegraphics[width=0.7\linewidth]{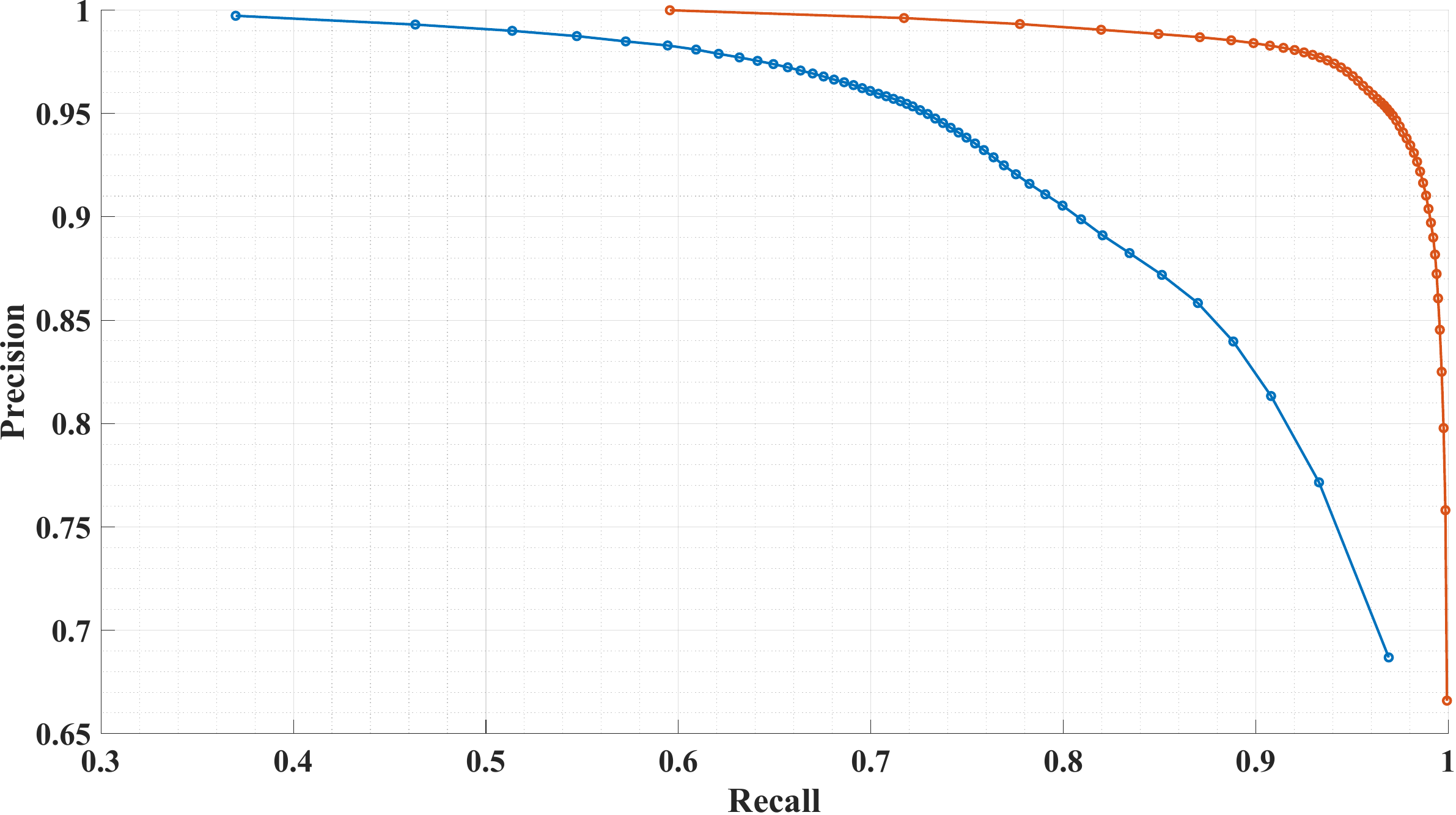}
            \caption{Floor segmentation results: orange curve (same environment) and blue curve (online dataset)}%The orange and blue curves correspond to results of the floor-segmentation from the same environment and from online dataset, correspondingly.}
            \label{fig:precision_versus recall}
        \end{figure}

    %In Module 1, Step 3, to train the proposed floor-segmentation neural net, approximately 100,000 data points are built. Data augmentation as used in building this data set is explained in Section \ref{section:methodologyA}. 
    In Module 1, Step 3, %the training was achieved in Google Colab\texttrademark{}. 
    The developed floor-segmentation method is tested using two datasets. The first dataset is collected in an indoor environment similar to the training set. The second dataset is collected from online databases. In order to collect these images, the keywords "hospital corridor", "office corridor", and "airport corridor" are used in a Google search. From the recommended images by this search engine, 33 images are selected. In selecting the images, a user judges the similarity of the images to the training set and selects more similar images.Some examples of the images and segmented floor are shown in \ref{fig:floor_segment_examples}
.    
    In order to illustrate the performance of the trained network quantitatively, the mean and standard deviation of the precision and recall values are calculated. These quantities are defined as  
            $R = TP/(TP + FN)$, \label{eq:recall}
            $P = TP/(TP + FP)$,\label{eq:precision}
    where $TP, FP, FN$ denote true positives, false positives, and false negatives, respectively. 
    In fact, precision is the fraction of predicted floor pixels that are actually floor pixels; this shows how accurate the positive predictions are.
    Recall is the fraction of actual floor pixels that were correctly identified. This shows how complete the detection is.
    These precision and recall values are measured at the pixel level. In order to achieve this, the pixels in the reference and segmented masks are compared to each other. The reference masks are acquired with the help of Grab-cut \cite{GrabCut2004} algorithm as implemented in MATLAB. % (this algorithm requires user interactions). 
    The recall and precision values of two test datasets are shown in Table \ref{tab:segmentation_results}. %Example images with the segmented floor (shown as green areas) are depicted in Fig. \ref{fig:floor_segment_examples}.
    The results for local images (images taken from the same environment as the training set) show a high precision and recall value of 0.96. The standard deviation is small and in the range of 0.05 to 0.07 for these values. For the images taken from online databases, while the precision stays high (0.96), the recall value drops to 0.71. Further, the standard deviations increase significantly up to 0.23 for the precision and 0.31 for recall. This indicates that more images from diverse environments are required to increase the accuracy of the floor-segmentation. In order to depict the precision-recall curve, we change the binary threshold of foreground (floor) and background segmentation. The corresponding precision-recall values are shown in Figure \ref{fig:precision_versus recall}.
     %   \CMB{Since this is a classification problem, how about plotting the Precision-Recall Curve?}

        \begin{table}[!htbp]
        \centering
        \caption{Performance comparison of tuned floor segmentation model (outcome of Module 1, Step 3)}
        \label{tab:segmentation_results}
        \renewcommand{\arraystretch}{1.2} % row spacing
        \begin{tabular}{l c c c c c}
        \hline
        Dataset         & Number of Images & Mean Precision & Std. Precision & Mean Recall  & Std. Recall \\
        \hline
        Local Images    & 31      &    0.96    &  0.05    & 0.96          & 0.07 \\
        Online Images   & 33      &      0.96   &  0.23   & 0.71         & 0.31 \\
        \hline
        \end{tabular}
        \end{table}

            \begin{figure}[t]
    \centering
    % Reduce spacing between subfigures
    \setlength{\tabcolsep}{1pt}
    \renewcommand{\arraystretch}{0}

    \begin{tabular}{ccc}
        \includegraphics[width=0.3\textwidth]{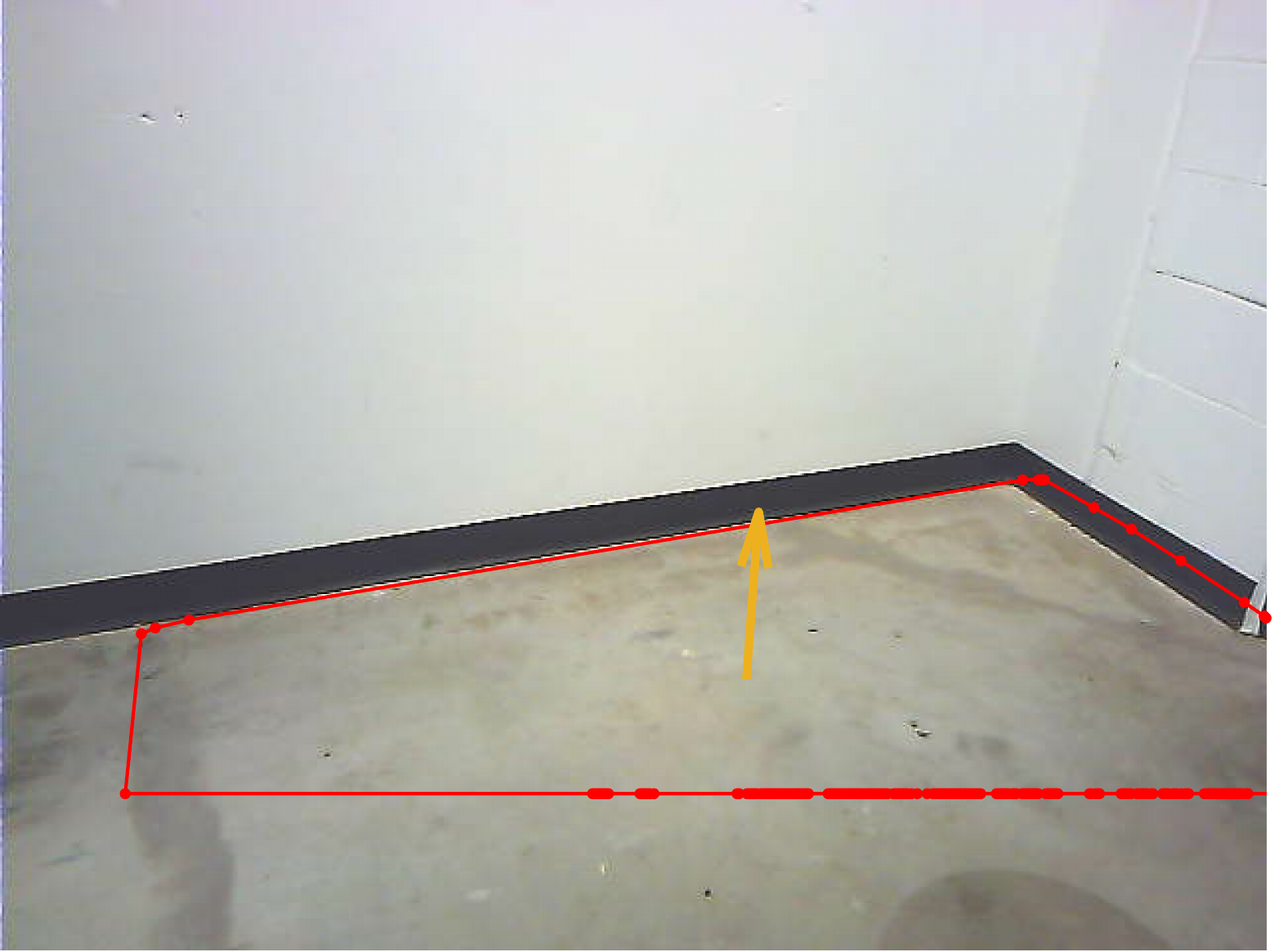} & 
        \includegraphics[width=0.3\textwidth]{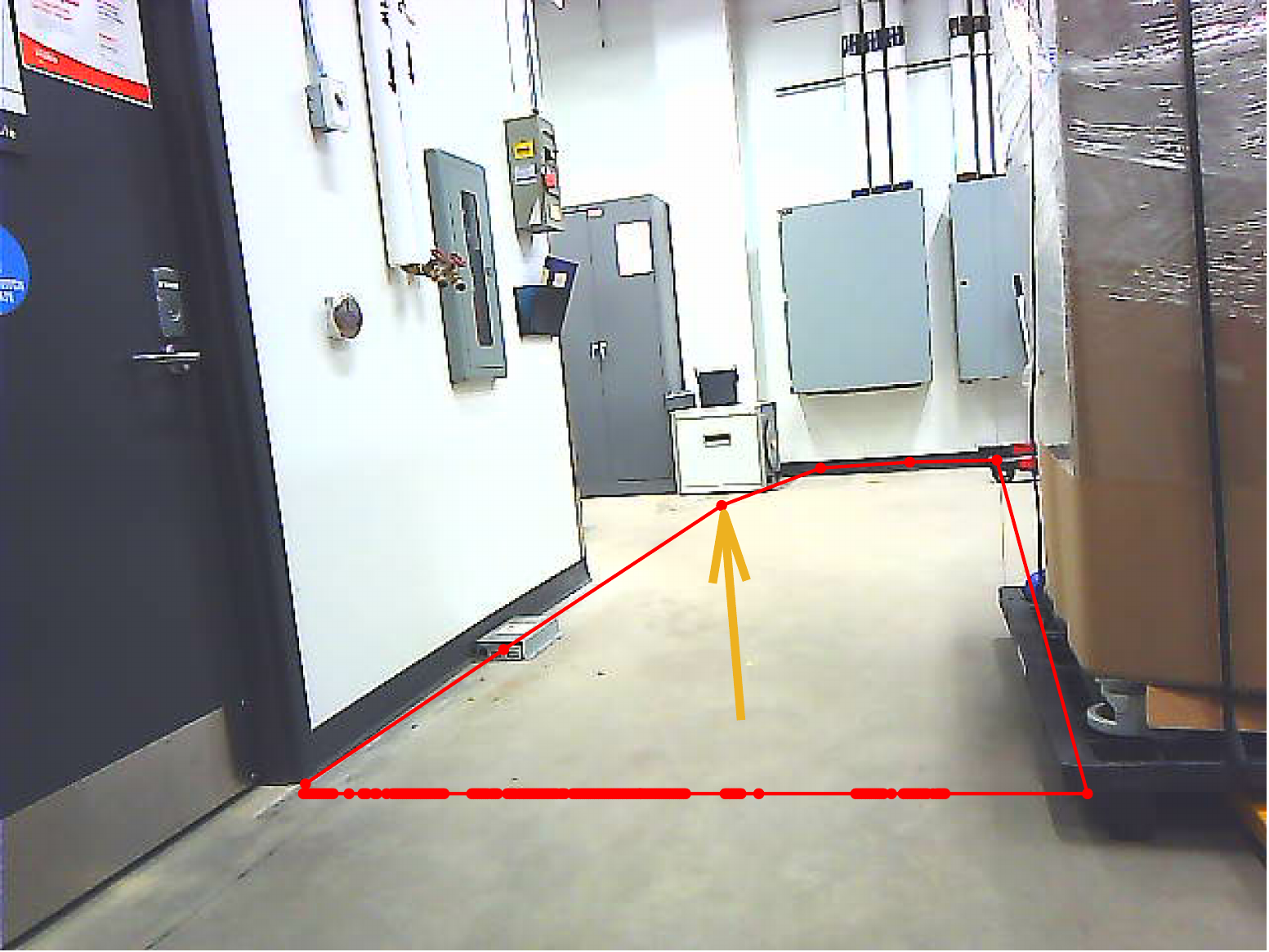}& 
        \includegraphics[width=0.3\textwidth]{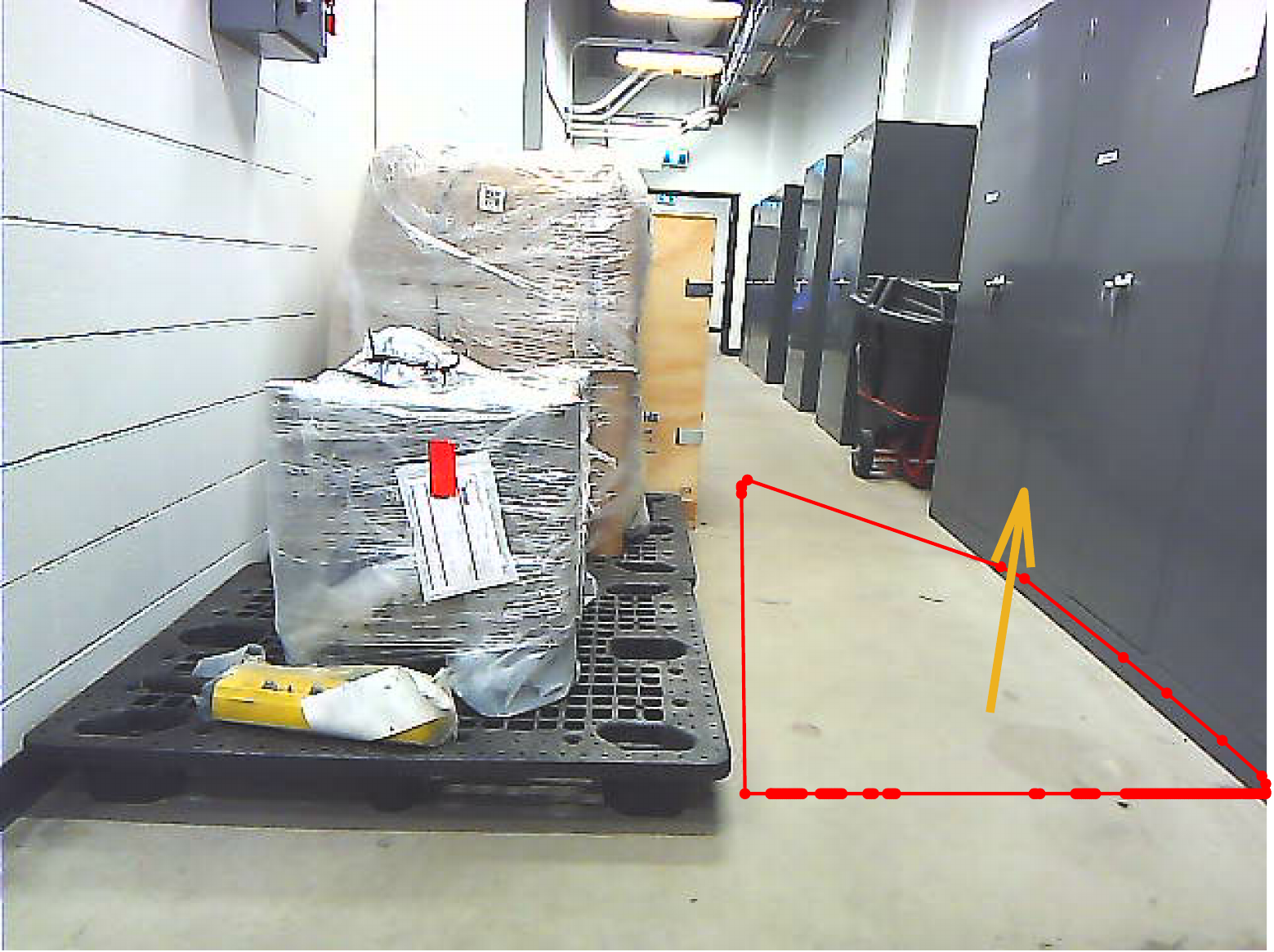} \\
        \includegraphics[width=0.3\textwidth]{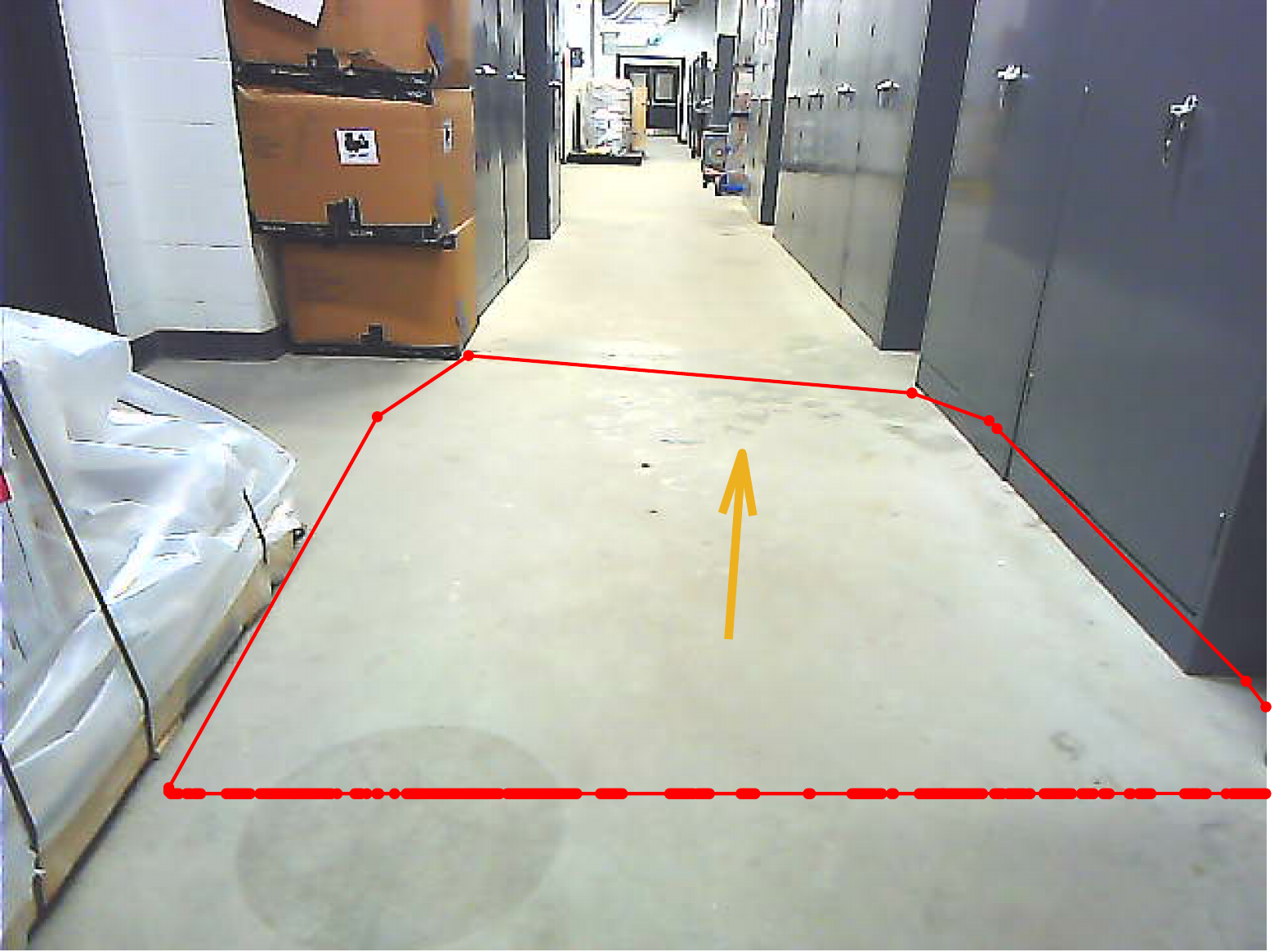} &
        \includegraphics[width=0.3\textwidth]{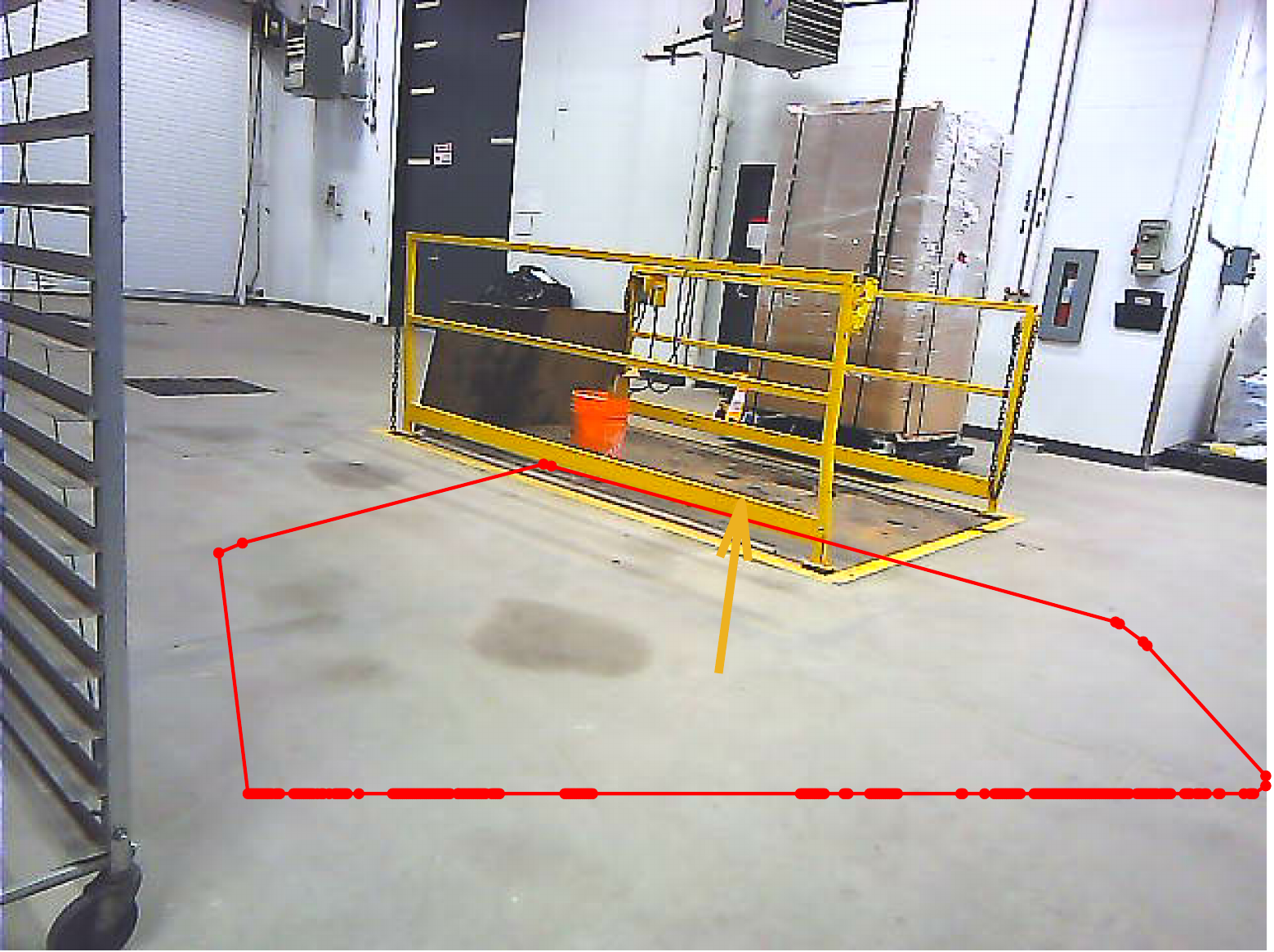} &
        \includegraphics[width=0.3\textwidth]{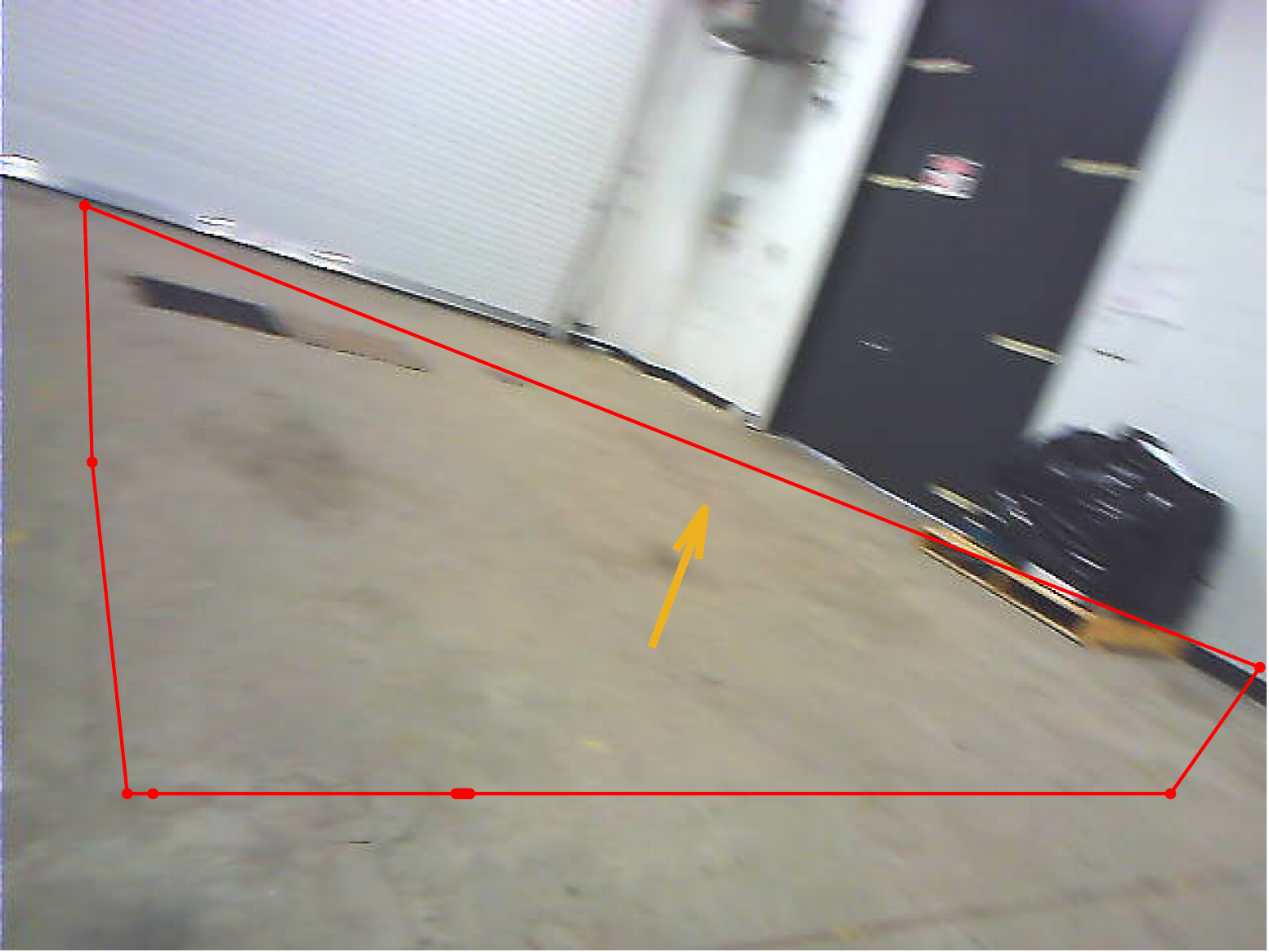}\\
    \end{tabular}

    \caption{Floor segmentation with surface normals in an industrial environment}%The floor-segment with normal}
    \label{fig:floor_segment_normal}
\end{figure}

       %TODO: Reporting module 2 is missing! the values from manufacturers,...

        %\CMB{Module 1, steps 4 and 5 results are missing} 
        The floor segment is back-projected to 3D space, where SVD-based plane fitting is achieved. The estimated plane normal is projected onto the image for visualization as shown in Figure \ref{fig:floor_segment_normal} (shown as orange arrow). Red polygons show the convex hull of the points in the depth channel that are on the floor segment and used to estimate the normal.
    
\begin{figure}[!b]
        \centering
        \includegraphics[width=1\textwidth]
        {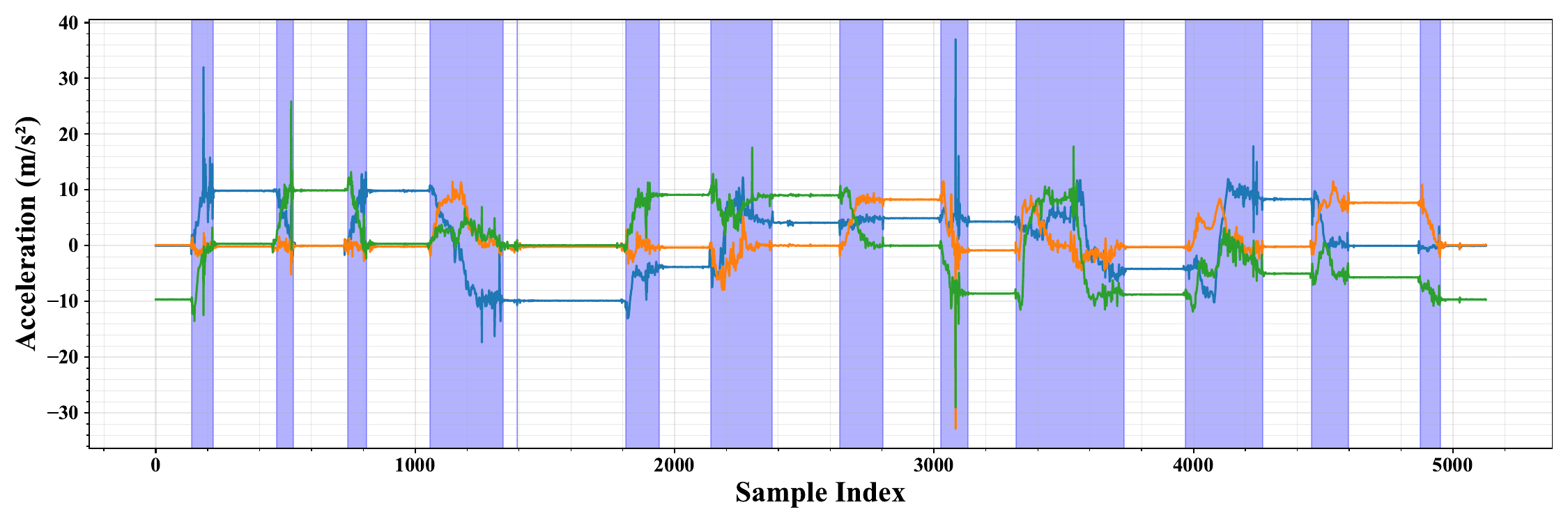}
        \caption{Three-axis accelerometer data (ISM330DHCX) with detected dynamic phases in purple}%Accelerometer data in three axes of ISM330DHCX, and the detected dynamic phases, highlighted in purple.}
        \label{fig:dynamic_phaes}
    \end{figure}

        After filtering the acceleration data in Module 2 Step 1, Step 2 involves static and dynamic phases that are necessary for estimating the intrinsic calibration parameters of the accelerometer. The detected dynamic phases are shown in Fig. \ref{fig:dynamic_phaes} in purple. The intervals in between dynamic phases, in addition to the initial and the final phases, mark the static phases. For brevity, the static phases detection for only one of the IMUs, specifically ISM330DHCX, is provided.

    The static/dynamic phases are utilized in Module 2, Step 3, to estimate the IMU intrinsic parameters. The final estimated values of these parameters are shown in Table \ref{tab:accelerometer_intrinsics}. The parameters that have the highest influence are biases and scale factors, as they constitute the largest portion of the error. For an ideal IMU, the calibration parameters for the scale values should be 1 in all three axes, and should be zero for the bias. In the next section, the importance of calibration for the MPU6050 is discussed. Based on Table \ref{tab:accelerometer_intrinsics}, MPU6050 exhibits much larger errors without calibration in both the scale and bias values, as this sensor belongs to an older generation of MEMS IMUs. %\CMB{how it is concluded that "MPU6050 exhibits much larger errors"} 
    
    \begin{table}[!htbp]
        \centering
        \caption{Estimated intrinsic calibration parameters of the IMUs}
        \begin{tabular}{c|c|c|c}
        \hline
        Parameter & ISM330DHCX & MPU6050 &  LSM6DSOX\\
        \hline
        $s_x$    & 0.9940 & 1.1335 & 0.9957\\
        $n_{xy}$ & 0.0186 & 0.3782 & 0.0081\\
        $s_y$    & 1.0131  & 0.9200 & 0.9703\\
        $n_{xz}$ & -0.0070 & 0.3140 & -0.0059\\
        $n_{yz}$ & -0.0159 & -0.1310 & 0.0028\\
        $s_z$    & 1.0023  & 0.9050 & 0.9997\\
        $b_x$    & 0.0225 & 0.7308 & -0.0128\\
        $b_y$    & 0.0086 & -0.5024 & -0.0108\\
        $b_z$    & -0.1313 & 1.6950 & -0.1189\\
        \hline
        \end{tabular}
        \label{tab:accelerometer_intrinsics}
    \end{table}   
       
    In order to test the extrinsic calibration algorithm (Module 3), the images of the test dataset was not included in the floor segmentation training to correctly assess the generalizability  of the trained neural nets to unseen scenes. The test set includes a variety of scenes. These include hallways, corners, and areas without any outstanding structures. 
    
    The accuracy of the proposed extrinsic calibration algorithm is evaluated using MATLAB's method \cite{matlabCameraIMUCalibration} and Kalibr \cite{rehder2016extending}. MATLAB's RGB-D/IMU extrinsic calibration is a target-based method that requires a checkerboard. The method first estimates the trajectories of a camera and an IMU, independently and in a subsequent step, estimates the unknown extrinsic calibration parameters. While estimation of the camera's trajectory with the help of a checkerboard can be achieved, such trajectory estimation for a low-cost MEMS IMU with 6 DoF does not provide a solution due to error accumulation. In order to address this challenge, the unknown extrinsic orientation parameters are calculated using the gyroscope measurement by modifying MATLAB's RGB/IMU extrinsic calibration method. In this modification, the angular velocities of the camera are calculated using the estimated trajectory. With the help of estimated angular velocities in the camera's frame and known gyroscope angular rate measurements in the IMU's body frame, the extrinsic calibration parameters are estimated using a variety of solutions to Wahba's problem. In particular, we have utilized the proposed RANSAC-based approach (as explained in Section \ref{section:methodology}) as well as SVD \cite{farrell1966least}, QUEST \cite{cheng2014improvement_quest}, and FLAE \cite{wu2017fast_quat}. 
    %\CMB{in the table, you name them by MATLAB. Do you mean you used Matlab functions?}%Ilyar: There are  two algorithms here.  One is the MATLAB extrinsic calibration toolbox. This calibration toolbox is modified by me to work with the  Wahba solver. MATLAB here refers to that extrinsic calibration toolbox. and MATLAB-SVD, for instance, refers to the fact that we estimate the trajectory using MATLAB toolbox and, in the end, apply SVD to solve Wahba's problem. I clarified this point in the text now. Please find it in a couple of paragraphs below, where I introduce the Table.
    SVD-based solution to Wahba's problem is an earlier approach, and it avoids representing the attitude with Euler or quaternion parameterizations; it directly estimates the rotation matrix. SVD is an alternative technique that is computationally expensive for real-time attitude estimation applications. Alternatively, QUEST utilizes quaternions and does not require a complete eigen vector decomposition, thus making it more suitable for online applications. FLAE is a non-iterative approach that also utilizes quaternions to estimate the attitude.
    The proposed method is further compared to Kablir \cite{rehder2016extending}. Kalibr is a well-known target-based calibration technique of IMU and camera, relying on targets. Unlike  MATLAB's extrinsic calibration toolbox, Kalibr estimates the unknown extrinsic calibration while achieving sensor fusion during the motion of the sensors. This method is widely recognized as the standard approach to assessing the extrinsic calibration methods. 
    
    In order to measure the accuracy of the proposed method (Module 3), geodetic distance is utilized. The distance between two rotation matrices is calculated as
        \begin{equation} \label{geodetic distance}
        d(\boldsymbol{R}_{\text{Kalibr}}, \boldsymbol{R}_{\text{ref}}) \;=\; \cos^{-1}\!\left(\frac{\operatorname{trace}\!\left(\boldsymbol{R}_{\text{Kalibr}}^{T} \boldsymbol{R}_{\text{ref}}\right) - 1}{2}\right).
    \end{equation}

    %\CMB{Introduce formula to calculate calibration error and describe it}

\begin{figure}
        \centering
        \includegraphics[width=1\linewidth]{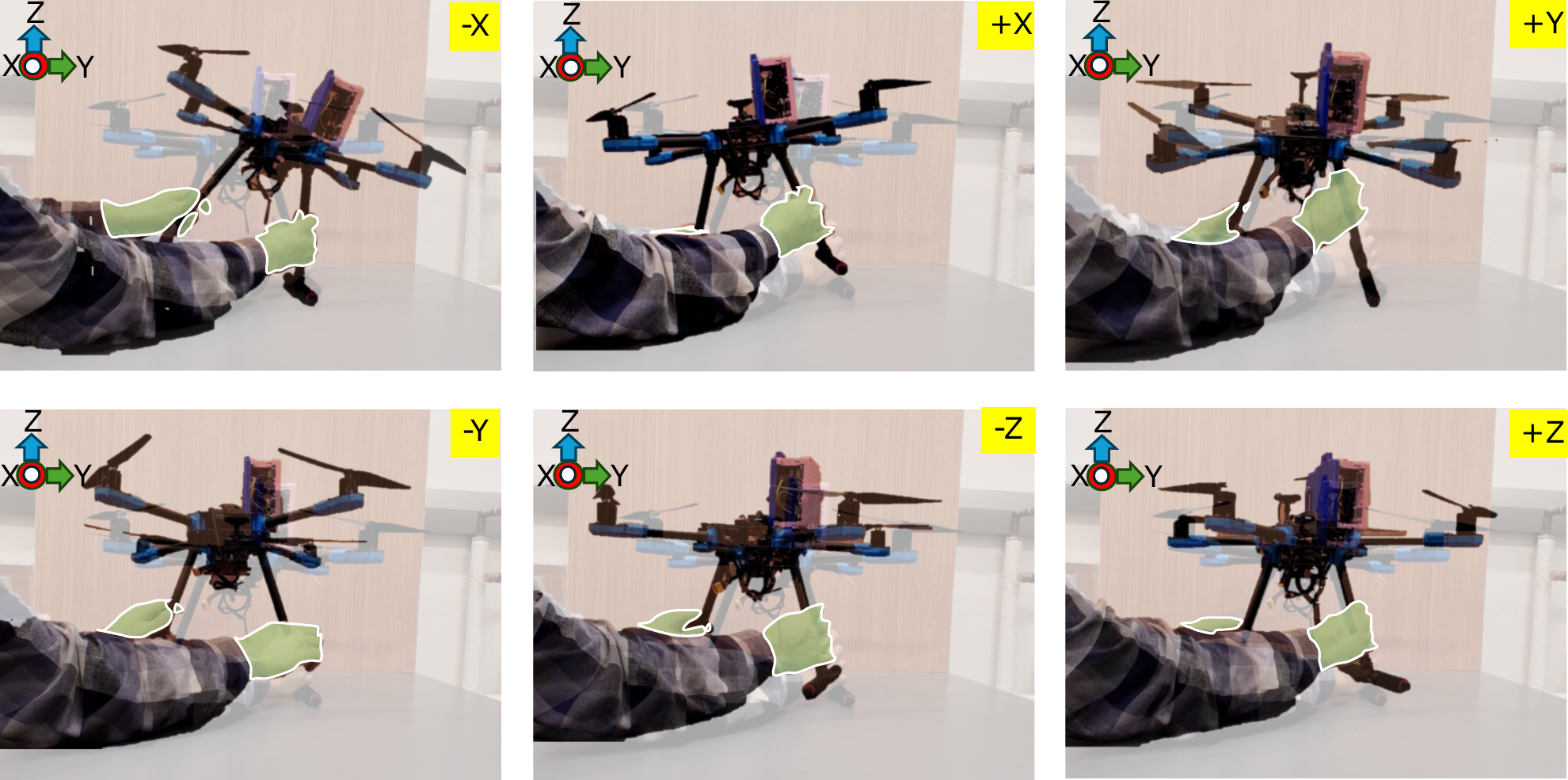}
        \caption{The sequence of rotations in $x$, $y$, and $z$ axes.}
        \label{fig:drone_axis_excitation}
    \end{figure}

    To ensure that all extrinsic orientation parameters are observable, the three sensor axes must be excited during the data collection process. To achieve this, the sequence of motions shown in Figure \ref{fig:drone_axis_excitation} is performed. In this figure, the drone (with the IMU/RGB-D attached) is first rolled forward and backward, exciting motion along the $x$-axis. Next, the drone is pitched, producing rotation about the $y$-axis. Finally, the drone is rotated about the $z$-axis to generate yaw motion. During this motion, the floor segment should be in the field of view of the RGB-D camera.
    
\begin{figure}
        \centering
        \includegraphics[width=1\linewidth]{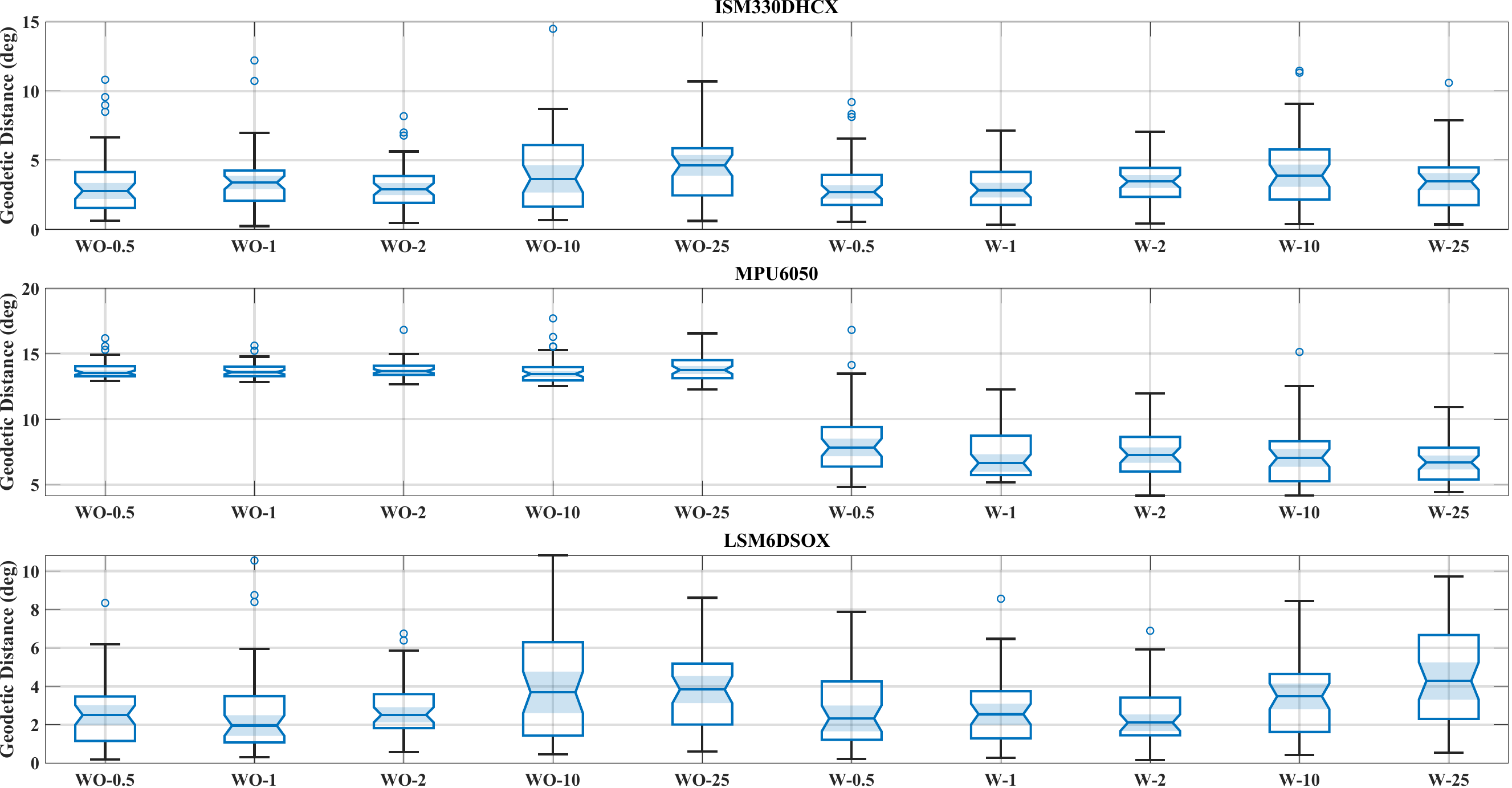}
        \caption{Comparison of extrinsic calibration errors for RGB-D-IMU using different Butterworth filter cut-off frequencies}%{Boxplot of the error of proposed and tested methods for extrinsic calibration of RGB-D and IMU. The results for different cut-off frequencies for Butterworth filtering are shown in this plot.}
        \label{fig:extrinsic_cutoff_comparison}
    \end{figure}

    We investigated the importance of Module 1, Step 1 on the accuracy of the extrinsic calibration. The extrinsic calibration accuracy is compared to KALIBR for cut-off frequency values of $0.5, 1, 2, 10,$ and $25$ Hz for the three sensors, as shown in Fig. \ref{fig:extrinsic_cutoff_comparison}. The trials are repeated 50 times. The results in this plot indicate that, if other factors are held constant (such as intrinsic and extrinsic calibration), the filtering cut-off frequency has only a small effect on the final extrinsic calibration accuracy. In particular, for ISM330DHCX and LSM6DSOX, lower cut-off frequencies provide slightly better results in terms of average accuracy. However, for the MPU6050, a higher cut-off frequency of 20 Hz yields the best results. For the subsequent steps, only the best cut-off frequency values are selected for comparison.

    Error comparison of robust methods for extrinsic orientation matrix estimation is shown in Table \ref{tab:imu_quartiles}. In this table, the results for the RANSAC-based version of the proposed is shown. Due to larger number of outliers in the data, other robust attitude estimators, including FLAE, QUEST, could not estimate the attitude. However, for MATLAB's checkerboard-based method, attitude estimation methods, RANSAC, SVD, QUEST, or FLAE, were able to estimate the calibration parameters and are included in this table. As we utilize different approaches to estimate the extrinsic orientation, given the trajectory from MATLAB's Extrinsic Calibration Toolbox, the MATLAB–Wahba solver naming convention is used in Table \ref{tab:imu_quartiles}, with the Wahba solver replaced by RANSAC, SVD, QUEST, or FLAE.
    % \EMB{The proposed\_RANSAC\_WO and proposed\_RANSAC\_W refers to ....
    % The MATLAB\_RANSAC, MATLAB\_QUEST, and MATLAB\_FLAE refers to ....
    % KALIBR refers to ....}

\begin{figure}[!t]
        \centering
        \includegraphics[width=1\linewidth]{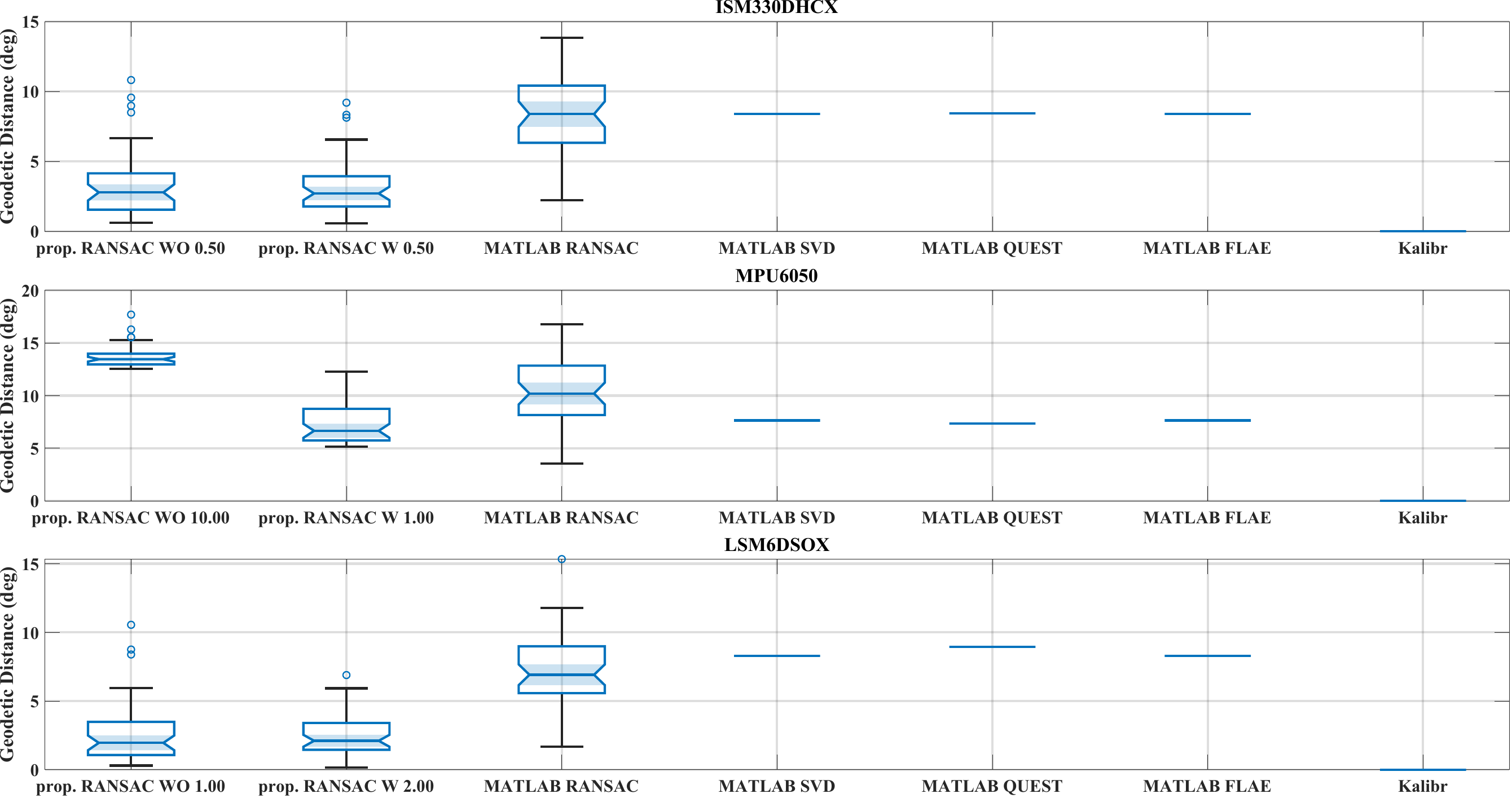}
        \caption{Comparison of RGB-D-IMU extrinsic calibration errors: proposed method vs. MATLAB's calibration}%Boxplot of the error of proposed and tested methods for extrinsic calibration of RGB-D and IMU. The proposed method is outperforming MATLAB's calibration method}
        \label{fig:boxplots}
    \end{figure}

    Table \ref{tab:imu_quartiles} illustrates the results with and without calibrating the accelerometer (Module 2, Step 3). 
    Since RANSAC is a non-deterministic algorithm, the  trials are repeated 50 times. The results of these 50 experiments are summarized as $25\%$, $50\%$ and $75\%$ quantiles. The methods such as SVD, QUEST, FLAE, and Kalibr are all deterministic methods, and thus they achieve the same values for these quantiles across the experiment (the results are also graphically summarized in Fig. \ref{fig:boxplots}).  
    
    The results in Table \ref{tab:imu_quartiles} illustrate that the proposed method outperforms MATLAB's algorithm. Further, the intrinsic calibration of the IMU has reduced the errors with a total error  $4.23^{\circ}$ achieved using RANSAC and intrinsic calibration (proposed method with calibration). In fact, RANSAC was the only method that results in a successful extrinsic calibration using the proposed method.

    \begin{table}[!htbp]
    \centering
    \caption{Error comparison of robust methods for extrinsic orientation matrix estimation}%Comparison of error (in degrees) of the robust techniques to estimate the extrinsic orientation matrix}
    \renewcommand{\arraystretch}{1.2} % row spacing
\begin{tabular}{l|ccc|ccc|ccc|c}
    \hline
    & \multicolumn{10}{c}{\textbf{Error in extrinsic rotation (degrees)}} \\
    \cline{2-11}
    \multirow{3}{*}{Method} 
      & \multicolumn{3}{c|}{ISM330DHCX} 
      & \multicolumn{3}{c|}{MPU6050} 
      & \multicolumn{3}{c|}{LSM6DSOX} & 
      \multirow{2}{*}{average} \\
      & 25\% & 50\% & 75\% & 25\% & 50\% & 75\% & 25\% & 50\% & 75\% &  \\
    \hline
        proposed\_RANSAC\_WO & \textbf{{1.56}} & 4.08 & 5.52 & 13.01 & 13.67 & 14.57 & \textbf{{1.48}} & \textbf{{3.12}} & \textbf{{4.62}} & 6.97 \\
        proposed\_RANSAC\_W  & 2.33 & \textbf{{3.65}} & \textbf{{5.38}} & \textbf{{4.62}} & \textbf{{5.20}} & \textbf{{6.33}} & 2.04 & 3.85 & 5.52 & \textbf{{4.23}}\\
        MATLAB\_RANSAC       & 6.49 & 7.97 & 9.93 &  7.95 &  9.77 & 11.78 & 4.64 & 6.70 & 8.93 & 8.15\\
        MATLAB\_SVD          & 8.40 & 8.40 & 8.40 &  7.64 &  7.64 &  7.64 & 8.28 & 8.28 & 8.28 & 8.11\\
        MATLAB\_QUEST        & 8.43 & 8.43 & 8.43 &  7.35 &  7.35 &  7.35 & 8.94 & 8.94 & 8.94 & 8.24 \\
        MATLAB\_FLAE         & 8.40 & 8.40 & 8.40 &  7.64 &  7.64 &  7.64 & 8.28 & 8.28 & 8.28 & 8.11\\
        KALIBR               & 0.00 & 0.00 & 0.00 &  0.00 &  0.00 &  0.00 & 0.00 & 0.00 & 0.00 & 0.001\\
    \hline
\end{tabular}
\label{tab:imu_quartiles}
\end{table}

    RANSAC parameters (such as outliers and the threshold for the distance) as shown in Algorithm \ref{alg:RANSAC} are summarized in Table \ref{tab:alg1_parameters}. The number of points to fit a RANSAC model is kept as three. This is the minimum number of points required to estimate extrinsic orientation parameters with the help of the developed approach, as explained in Section \ref{section:methodology}. The outlier rejection threshold is set to the values in Table \ref{tab:alg1_parameters} after trial and error. The detected inlier-to-outlier ratios are shown for each sensor in this table as well. As with the calibration parameters, it is expected to have a lower inlier to outlier ratio for MPU6050 as this sensor belongs to a relatively older generation compared to ISM33DHCX and LSM6DSOX.

    \begin{table}[!htbp]
        \centering
        \caption{Effect of threshold selection on proposed method performance}%Proposed method performance with different thresholds}
        \label{tab:alg1_parameters}
        \renewcommand{\arraystretch}{1.2}
        \begin{tabular}{l c c c}
        \hline
        Configuration & ISM330DHCX & MPU6050 & LSM6DSOX \\
        \hline
        model points = 3, outlier rejection threshold = 0.035 
            & 0.24 & 0.22 & 0.22 \\
        model points = 3, outlier rejection threshold = 0.3   
            & 0.31 & 0.22 & 0.32 \\
        \hline
    \end{tabular}
    \end{table}

    % \CMB{Is it possible to plot components of Equation \ref{eq:error term} and error vs time in the resits section, for IMU and camera?}

\section{Conclusions and Future Work}\label{section:conclusions}

In this study, we proposed a target-less approach to estimate the extrinsic calibration parameters of an IMU and an RGB-D camera with the help of a deep neural net-based floor-segmentation. The proposed method utilized segmentation to extract the pixels corresponding to the floor. The corresponding pixels in the depth image are utilized to estimate the floor-normal. This vector and the sensed gravity vector in the accelerometer's frame are utilized to estimate the extrinsic calibration parameters using RANSAC.

The developed method is tested using Kalibr as the reference solution and MATLAB's calibration techniques. Further, RANSAC-based estimation is compared to SVD, FLAE, and other robust estimators. The developed method is shown to be able to achieve a similar level of accuracy of the extrinsic calibration without relying on any targets. Further, it is shown that RANSAC-based estimation is most suitable for the developed approach due to its capacity to reject outliers in the data, while other methods, such as FLAE, failed. It is shown that intrinsic calibration of IMU's plays a critical role in improving the overall accuracy. 

The developed floor-segmentation is tested with two datasets. The precision of the developed method is reported to achieve approximately 96.0 precision and recall values for a test data set similar the training data set. For the outdoor environment, the mean precision value remains high, while the recall drops to approximately 71.0 percent. The developed floor-segmentation, however exhibits errors under low-illumination conditions and high levels of reflection on the floor and scenes. Thus, in the future, a model with wider applicability for the floor-segmentation can be developed. 

%\CMB{Limitations section: Briefly discuss when your method might fail (e.g., cluttered floors, outdoor uneven terrain).}
One of the key limitations of the proposed method is that it assumes the floor normal vector is parallel to the gravity vector. Although this assumption is almost always valid in indoor, human-made environments, it often does not hold in outdoor environments. Furthermore, the proposed method cannot be applied if the floor segment is completely obscured by clutter.
%\CMB{Future work: Mention possible extensions (e.g., adapting to outdoor environments, real-time implementation).}
In the future, we plan to extend the proposed method to outdoor scenes. In particular, in cities with tall buildings (commonly referred to as the Manhattan World), the structural regularity can be leveraged to estimate the gravity direction from the camera, thereby enabling extrinsic calibration of the camera and IMU as well.

\section*{Funding Sources}
This work was supported by the Natural Sciences and Engineering Research Council of Canada (NSERC), the Government of Alberta, Alberta Innovates, and the Schulich School of Engineering at the University of Calgary. Funding was awarded to Dr. Mahdis Bisheban, Director of the Intelligent Dynamics and Control Lab and Assistant Professor at the University of Calgary.

\section*{Code Availability}
The source code and all calibration scripts used in this work are available at \href{https://github.com/IlyarAbadi/RGBD2IMU_Calibration}{RGBD2IMU\_Calibration} Github repository.

\bibliography{sample}
	
\end{document}